\documentclass[nohyperref]{article}

\usepackage{microtype}
\usepackage{graphicx}
\usepackage{subfigure}
\usepackage{booktabs} 

\usepackage{hyperref}

\usepackage[accepted]{icml2023}

\usepackage{amsmath}
\usepackage{amssymb}
\usepackage{mathtools}
\usepackage{amsthm}
\usepackage[hang,flushmargin]{footmisc}

\usepackage[capitalize,noabbrev]{cleveref}

\theoremstyle{plain}

\theoremstyle{definition}

\theoremstyle{remark}

\usepackage[textsize=tiny]{todonotes}

\usepackage[table,xcdraw]{xcolor}
\usepackage{cuted}
\usepackage{pifont}
\usepackage{lipsum}
\usepackage{wrapfig}
\usepackage{overpic}
\usepackage{overpic}          
\usepackage{capt-of}
\usepackage{multirow}
\usepackage{booktabs}
\usepackage{comment}
\usepackage{comment}
\usepackage{amsmath}
\usepackage{amssymb}
\usepackage{adjustbox}
\usepackage{soul,color}
\usepackage{enumitem} 
\definecolor{turquoise}{cmyk}{0.65,0,0.1,0.3}
\definecolor{purple}{rgb}{0.65,0,0.65}
\definecolor{dark_green}{rgb}{0, 0.5, 0}
\definecolor{orange}{rgb}{0.8, 0.6, 0.2}
\definecolor{red}{rgb}{0.8, 0.2, 0.2}
\definecolor{darkred}{rgb}{0.6, 0.1, 0.05}
\definecolor{blueish}{rgb}{0.0, 0.3, .6}
\definecolor{light_gray}{rgb}{0.7, 0.7, .7}
\definecolor{pink}{rgb}{1, 0, 1}
\definecolor{greyblue}{rgb}{0.25, 0.25, 1}
\newcommand{\real}{\mathbb{R}}

\usepackage{blindtext}

\renewcommand{\paragraph}[1]{\vspace{1em}\noindent\textbf{#1}.}

\newcommand{\xmark}{\ding{55}}%

\newcommand{\best}[1]{{\color[HTML]{3531FF}{\textbf{#1}}}}
\newcommand{\sbest}[1]{{\color[HTML]{FE0000}{\textit{#1}}}}
\newcommand{\mailtodomain}[1]{\href{mailto:#1@robots.ox.ac.uk}{\nolinkurl{#1}}}

\icmltitlerunning{~\hfill
Multi-Modal Classifiers for Open-Vocabulary Object Detection 
\hfill \thepage}
\begin{document}

\twocolumn[
\icmltitle{Multi-Modal Classifiers for Open-Vocabulary Object Detection}



\begin{icmlauthorlist}
\icmlauthor{Prannay Kaul}{vgg}
\icmlauthor{Weidi Xie}{vgg,sjt,sjai}
\icmlauthor{Andrew Zisserman}{vgg}
\end{icmlauthorlist}

\icmlaffiliation{vgg}{Visual Geometry Group, University of Oxford}
\icmlaffiliation{sjt}{CMIC, Shanghai Jiao Tong University}
\icmlaffiliation{sjai}{Shanghai AI Lab}

\icmlcorrespondingauthor{Prannay Kaul}{prannay@robots.ox.ac.uk}

\centering{\url{https://www.robots.ox.ac.uk/vgg/research/mm-ovod/}}

\vskip 0.2in
]



\printAffiliationsAndNotice{}  

\begin{abstract}
The goal of this paper is open-vocabulary object detection~(OVOD)
--- building a model that can detect objects beyond the set of categories seen at training,
thus enabling the user to specify categories of interest at inference without the need for model retraining.  
We adopt a standard two-stage object detector architecture, 
and explore three ways for specifying novel categories: 
via language descriptions, via image exemplars, or via a combination of the two.
We make three contributions:
{\em first}, we prompt a large language model~(LLM) to generate informative language descriptions for object classes,
and construct powerful text-based classifiers;
{\em second}, we employ a visual aggregator on image exemplars that can ingest any number of images as input,
forming vision-based classifiers; and
{\em third}, we provide a simple method to fuse information from language descriptions and image exemplars,
yielding a multi-modal classifier.
When evaluating on the challenging LVIS open-vocabulary benchmark we demonstrate that:
(i) our text-based classifiers outperform all previous OVOD works;
(ii) our vision-based classifiers perform as well as text-based classifiers in prior work; (iii) using multi-modal classifiers perform better than either modality alone; and finally, (iv) our text-based and multi-modal classifiers yield better performance than a fully-supervised detector.
\end{abstract}

\section{Introduction}\label{sec:intro}
\vspace{-0.1cm}
In this paper, we consider the problem of open-vocabulary object detection~(OVOD),
which aims to localise and classify visual objects beyond the categories seen at training time. 
One may consider its usefulness from the perspective of online inference,
when users want to freely specify categories of interest at inference time without
the need or ability to re-train models. 
To specify categories of interest, three obvious ways exist, namely:
(1) text-based, \eg~name the category or describe it in text form;
(2) vision-based, \eg~give image examples;
(3) multi-modal, \eg~indicate the category jointly with text and image.

\begin{figure}[t]
\scriptsize
\includegraphics[width=1.0\linewidth]{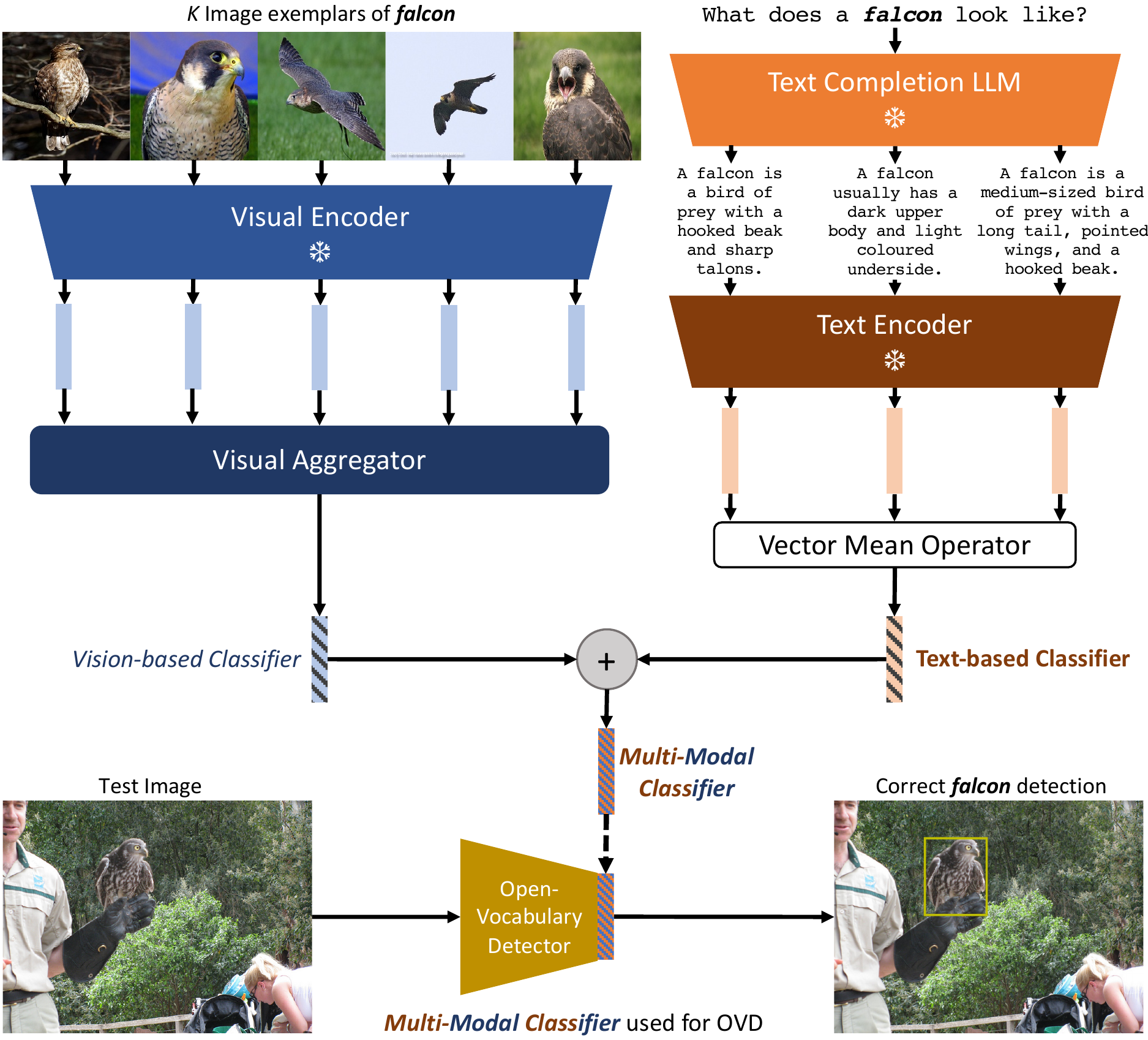}
\vspace{-0.4cm}
\caption{
Overview of the  architecture for generating
text-based, vision-based, or multi-modal classifiers for OVOD.
Vision (top left): A frozen visual encoder ingests image exemplars of {\bf falcon}
producing an embedding per exemplar.
A trained aggregator takes these embeddings as input and produces a
\emph{vision-based classifier}.
Text (top right): A text completion LLM is prompted to give descriptions of
a {\bf falcon} which are then encoded by a text encoder and averaged yielding
a \emph{text-based classifier}.
Multi-Modal (middle): \emph{Multi-Modal classifiers} are generated by adding the
vision-based and text-based classifiers together.
OVOD (bottom): The multi-modal classifier is used to detect the {\bf falcon} in a
standard model.
Note, all three types of classifier: vision-based, text-based and multi-modal,
can be used on the detector head for OVOD.}\label{fig:teaser}
\vspace{-0.4cm}
\end{figure}


Existing works~\cite{Bansal2018,Gu2021,Zareian2021,Zhou2022,Feng2022} have
explored replacing the learnt classifiers in a traditional detector with text embeddings, 
that are computed by passing the class name into a pre-trained text
encoder with manual prompts, such as ``a photo of a dalmatian'',
however, this design can be sub-optimal from three aspects:
{\em first}, the discriminative power of generated text embeddings relies
entirely on the internal representation of the pre-trained text encoder,
potentially leading to lexical ambiguities --- 
\eg~``nail'' can either refer to ``the hard surface on the tips of the fingers''
or ``a small metal spike with a flat tip hammered into wood to form a joint'' 
--- simply encoding the {class name} will not be able to distinguish the two concepts;
{\em second}, the {class name} for objects of interest may be unknown to the user,
while exemplar images can be easily acquired
--- \eg~``dugong'' refers to a herbivorous marine mammal with an adorable, 
plump appearance, a dolphin tail, round head and downward snout~(an example is in Section~\ref{sec:dugong} of the Appendix);
{\em third}, in cases where multi-modal information is preferable to specify the category of interest,
\eg~a species of butterfly with a distinctive wing pattern
--- language descriptions can be unsuitably long to capture all the intricacies of a given category,
while an exemplar image can ``tell a thousand words'' and act as an effective complement to text. 


To tackle these limitations, 
we propose a multi-modal open-vocabulary object detector,
with the classifier of the detector for a particular category being constructed
via natural language descriptions, image exemplars or a combination of the two.
Specifically, we establish an automatic approach for sourcing visual descriptions of the object categories,
by prompting a large language model with questions, \eg~``What does a dalmatian look like?'',
yielding ``A dalmatian is typically a large dog with a short coat of black spots on a white background''. 
Such a description provides additional visual cues to enhance the discriminative power of the classifier
generated from a text encoder.
For cases where collecting suitable informative language descriptions may be difficult
or require unnecessarily long descriptions to establish the differences between classes,
\eg~the dog breeds ``pug'' and ``bulldog'' have similar descriptions,
we can generate classifiers from image exemplars --- RGB images of the class of interest.
Finally, we suggest a simple method to fuse both language descriptions and image exemplars,
yielding multi-modal classifiers that perform better than either modality individually.

We explore the issue of how best to combine the set of language descriptions
and the set of image exemplars into a classifier,
by comparing the performance of aggregation methods on a standard detector architecture.
By evaluating on the challenging LVIS~\cite{Gupta2019} open-vocabulary object detection benchmark we show that:
(i) our automated method for sourcing rich natural language descriptions yields
text-based classifiers superior to those of previous work that rely entirely on the class name;
(ii) vision-based classifiers can be effectively constructed by a visual aggregator,
enabling novel categories to be detected by specifying image exemplars;
(iii)  natural language descriptions and image exemplars can be simply combined to produce multi-modal classifiers,
which perform better than either modality individually, and achieve superior results to existing approaches.
\newcommand{\parb}[1]{\par{\noindent \bf #1}}
\vspace{-0.2cm}
\section{Related Work}\label{sec:rw}
\vspace{-0.05cm}
\parb{Closed-Vocabulary Object Detection}
is one of the classical computer vision problems, making a full overview here
impossible.
Therefore, we outline some key milestones.
In general, modern object detection methods can be cast into two sets:
one-stage and two-(multi-)stage detectors.
\emph{One-stage} detectors directly classify and regress bounding boxes
by either densely classifying a set of predefined anchor
boxes~\cite{Redmon16,Redmon2018,Liu16,Lin2017a,Tan2020},
each which may contain an object, or densely searching for geometric entities of objects
\eg~corners, centre points or boxes~\cite{Law2018,Zhou2019,Tian2019}.
Conversely, most \emph{two-stage} detectors first propose class-agnostic
bounding boxes that are pooled to fixed size region-of-interest (RoI)
features and classified by a sub-network in the second stage~\cite{Girshick15,Ren16,Li2019}.
Two-stage detectors are extended to \emph{multi-stage} detectors in which
the additional stages refine predictions made by the previous
stage~\cite{Cai2018a,Chen2019c,Zhou2021}.
A unique work in this area is that of~\cite{Carion2020} which uses the
Transformer architecture~\cite{Vaswani17} to treat object detection as a set
prediction problem.
{\bf Note that}, the classifiers in these object detectors are jointly learnt on a training set,
therefore only objects seen at training time can be detected during inference time,
thus termed closed-vocabulary object detection.

\vspace{-0.05cm}
\parb{Open-Vocabulary Object Detection} goes beyond closed-vocabulary object
detection and enables users to expand/change the detector vocabulary at inference time,
without the need for model re-training.
Recently, OVOD has seen increased attention and progress primarily
driven by the emergence of large-scale vision-language models (VLMs) \eg~CLIP
and ALIGN~\cite{Radford2021,Jia2021}, which jointly learn image and natural
language representations.

ViLD~\cite{Gu2021} distills representations from VLMs.
First, groundtruth bounding boxes are used to crop an image and an
embedding for the box is sourced from a frozen VLM image encoder.
An object detection model is learnt by matching overlapping region-of-interest
(RoI) features with the embedding for the relevant box from
the VLM image encoder using a L$1$ reconstruction loss.
RegionCLIP~\cite{Zhong2022} uses image-caption data to construct region-wise
pseudo-labels, followed by region-text contrastive pre-training before transferring to detection.
GLIP and MDETR~\cite{Li2022,Kamath2021} use captions to cast detection as
a phrase grounding task and use early fusion between the image and
text modalities, increasing complexity.
OVR-CNN~\cite{Zareian2021} uses large image-caption data to pre-train a
detector to learn a semantic space and finetunes on smaller detection data.
OWL-ViT~\cite{Minderer2022} follows OVR-CNN but makes use
of large transformer models and even larger image-caption data.
OV-DETR~\cite{Zang2022} modifies the DETR framework for closed-vocabulary object detection~\cite{Carion2020}
to make it suitable for the open-vocabulary setting.
We note that OV-DETR can condition object detection on image exemplars,
but only provides some qualitative examples,
whereas we quantitatively benchmark our method using vision-based classifiers.
Detic and PromptDet~\cite{Zhou2022,Feng2022} improve open-vocabulary detection by
making use of image classification data to provide weak supervision on a large set of categories.
Our work uses Detic as a starting point in experiments,
and we investigate different methods for constructing the classifiers.



\vspace{-0.05cm}
\parb{Low-shot Object Detection.}
Despite garnering less attention in recent literature,
some low/few-shot object detection works make use of image-conditioned
object detection~\cite{Kang2019,Hsieh2019,Ssokin2020,Chen2021} in which image
exemplars of novel categories are encoded at inference time and used to detect novel category instances.
These works focus on architectural advances usually leveraging attention
between the novel class image exemplars and the inference time
image~\cite{Chen2021,Hsieh2019}.

There are an increasing number of low/few-shot object detection works
based on finetuning detector parameters on limited groundtruth data
for novel categories~\cite{Wang2020,Sun2021,Qiao2021a,Kaul22}.
These works finetune the detector on limited groundtruth novel instances
and so are not related to open-vocabulary object detection using vision-based
classifiers, where no novel instances are available for re-training/finetuning.

\vspace{-0.05cm}
\parb{Natural Language for Classification.}
Natural language is a rich source of semantic information for classification.
CLEVER~\cite{Choudhury21} matches descriptions of images in simple text form
with descriptions from expert databases \eg~Wikipedia to perform fine-grained
classification.
ZSLPP~\cite{Elhoseiny2017} extracts visual information from
large-scale text to identify parts of objects and perform zero-shot
classification.
The work by~\cite{Menon2022} uses class descriptions from GPT-3~\cite{Brown2020} for classification,
analysing which parts of the description contribute to classification decisions.
CuPL~\cite{Pratt2022} uses a GPT-3 model to provide detailed
descriptions enabling improved zero-shot image classification.
The learnings from this work inform our use of natural language
descriptions sourced from LLMs.

\newcommand{\mathset}[2]{\mathcal{#1}^\textsc{#2}}
\newcommand{\classifier}[1]{\mathbf{w}_{\textsc{#1}}^{c}}


\begin{figure*}[htb!]
\begin{minipage}[t]{0.52\linewidth}
\vspace{-6.73cm}
\scriptsize
\centering
\includegraphics[width=1.0\linewidth]{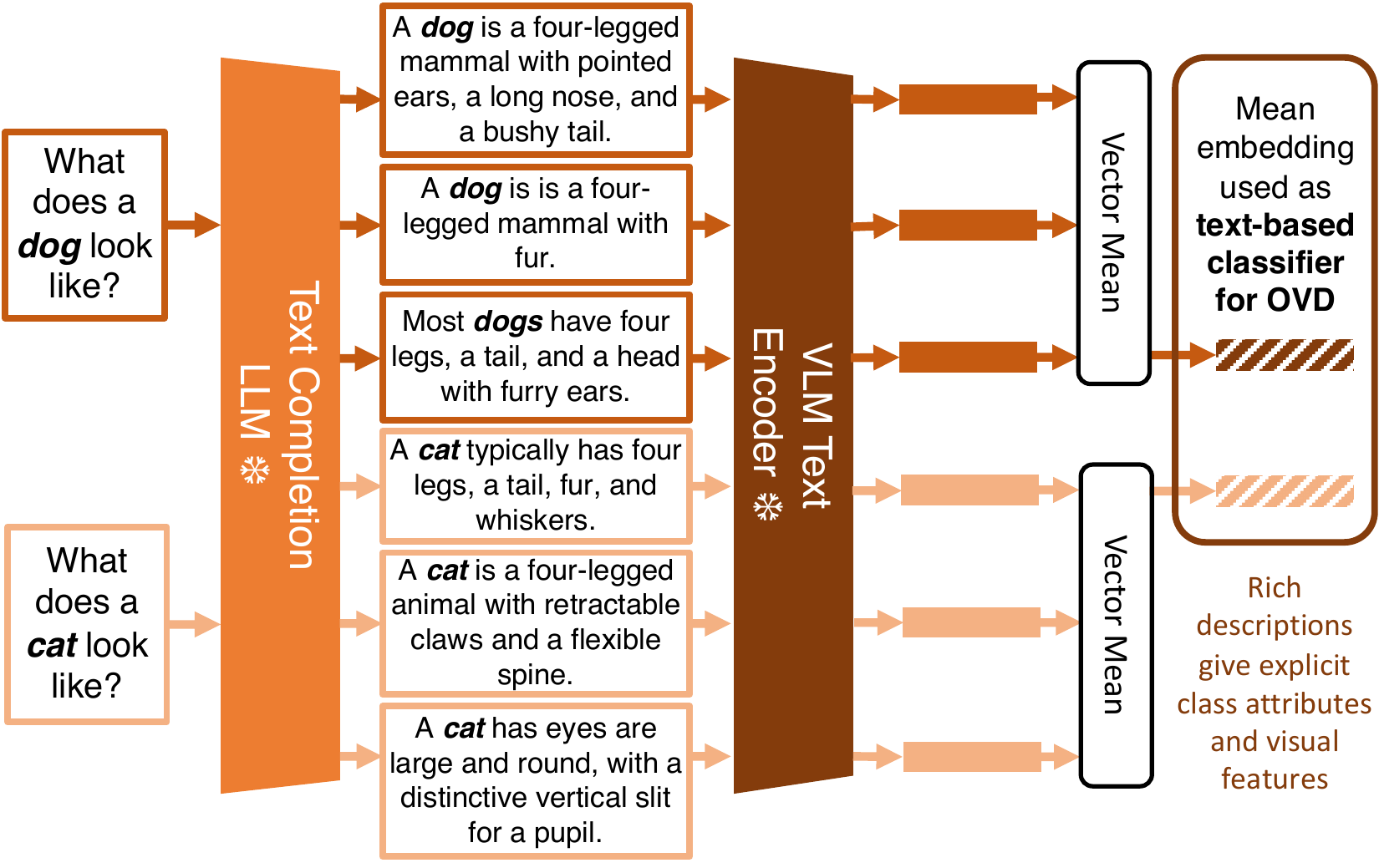}
\vspace{-0.5cm}
\caption{
Generating powerful text-based classifiers.
A LLM (GPT-3) is used to generate multiple rich descriptions of the class of
interest.
These descriptions are then encoded with the CLIP~\cite{Radford2021} VLM text
encoder.
The descriptions are more informative than the simple phrases,
such as ``(a photo of) a {\bf dog}'' or ``(a photo of) a {\bf cat}'',
used in previous work such as Detic and ViLD. 
Additional examples of class descriptions are given in the
Appendix (Section~\ref{sec:add_cls_desc}).}\label{fig:nl_emb}
\end{minipage}
\hfill
\begin{minipage}[t]{0.46\linewidth}
\scriptsize
\includegraphics[width=1.0\linewidth]{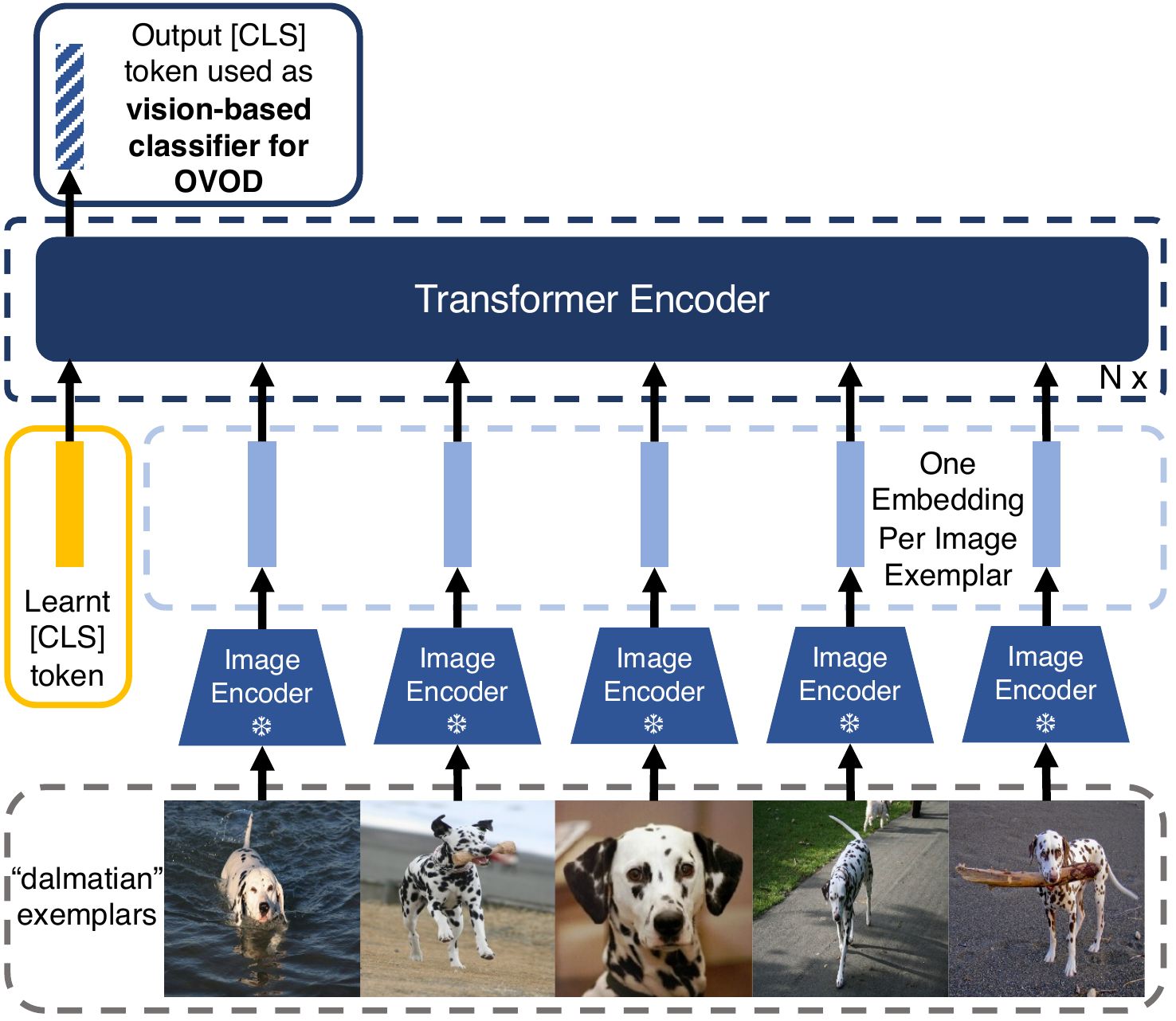}
\vspace{-0.47cm}
\caption{
Generating an OVOD vision-based classifier from  a set of  image exemplars.
A stack of transformer blocks is used to combine embeddings of multiple
exemplars belonging to the same category.}\label{fig:img_emb}
\end{minipage}
\vspace{-0.15cm}
\end{figure*}

\section{Method}\label{sec:meth}
\vspace{-0.1cm}
In this section, we start by providing background on open-vocabulary object detection~(OVOD),
then outline the proposed methods for constructing classifiers from
language descriptions of a category~(text-based classifiers, Section~\ref{ssec:nl_temp})
and image exemplars~(vision-based classifiers, Section~\ref{ssec:img_temp}).
Our final method combines the classifiers found from language descriptions and image exemplars,
yielding multi-modal classifiers~(Section~\ref{ssec:nl_img_temp}).


\vspace{-0.1cm}
\subsection{Preliminaries}\label{ssec:prelim}
\parb{Problem Scenario. }
Given an image~($\mathbf{I}~\in~\real^{3\times H\times W}$) fed input to an
open-vocabulary object detector, two outputs are generally produced:
(1) classification, in which a class label, $c_j \in \mathset{C}{test}$,
is assigned to the $j^{\text{th}}$ predicted object in the image,
and $\mathset{C}{test}$ refers to the category vocabulary desired at inference time;
(2) localisation, with bounding box coordinates, $\mathbf{b}_j \in \real^{4}$,
denoting the location of the $j^{\text{th}}$ predicted object.
In accordance with the setting introduced by Detic~\cite{Zhou2022},
two datasets are used at training time: a detection dataset, $\mathset{D}{det}$,
containing bounding box coordinates, class labels and associated images,
addressing a category vocabulary, $\mathset{C}{det}$;
and an image classification dataset, $\mathset{D}{img}$, 
containing images with class labels only,
addressing a category vocabulary, $\mathset{C}{img}$.
In the most general case there are no restrictions on the overlap or lack
thereof between the sets $\mathset{C}{test}, \mathset{C}{det}$ and $\mathset{C}{img}$.

\parb{Architecture Overview.} 
In this work, we make use of a popular multi-stage detector based on
CenterNet2~\cite{Zhou2021} as done in Detic.
This detector, with outputs $\left\{c_j, \mathbf{b}_j\right\}^{M}_{j=1}$,
can be formulated as (for simplicity we consider the two-stage variant below):
\begin{align}
&\left\{f_j\right\}^{M}_{j=1} = \label{eq:a}
\Phi_\textsc{roi}\circ\Phi_\textsc{pg}\circ
\Phi_\textsc{enc}\left(\mathbf{I}\right)  \\
&\left\{
\mathbf{b}_j, c_j \right\}^{M}_{j=1} = \label{eq:b}
\left\{
\Phi_\textsc{bbox}\left(f_j\right), 
\Phi_\textsc{cls}\circ\Phi_\textsc{proj}\left(f_j\right)
\right\}^{M}_{j=1} 
\vspace{-20pt}
\end{align}

where, each input image is first sequentially processed by a set of operations:
an image encoder~$\left(\Phi_\textsc{enc}\right)$;
a proposal generator~$\left(\Phi_\textsc{pg}\right)$;
a region-of-interest (RoI) feature pooling module~$\left(\Phi_\textsc{roi}\right)$,
yielding a set of RoI features, $\left\{f_j\right\}^{M}_{j=1}$.
The RoI features are processed by a bounding box module~$\left(\Phi_\textsc{bbox}\right)$ to infer position of objects, 
$\left\{\mathbf{b}_j\right\}^{M}_{j=1}$.
Additionally, the RoI features are processed by a classification module,
consisting of a linear projection~$\left(\Phi_\textsc{proj}\right)$,
and $C$ classification vectors or classifiers~$\left(\Phi_\textsc{cls}\right)$,
yielding a set of class labels, $\left\{c_j\right\}^{M}_{j=1}$
($C$ is the size of the category vocabulary).

In closed-vocabulary object detection all parameters listed above are learnt
during training on $\mathset{D}{det}$.
While in the open-vocabulary scenario, 
the classifiers~($\Phi_\textsc{cls}$) \emph{are not}
learnt during training but are instead generated separately
from an alternative source, \eg~a pre-trained text encoder.
This allows $\mathset{C}{test} \neq \mathset{C}{det}$,
as the classifiers, $\Phi_\textsc{cls}$,
for a specific set of user defined classes can be generated at inference time.
In the following sections, we describe different options for constructing such classifiers:
from natural language, from image exemplars, or from a combination of the two.

\vspace{-0.1cm}
\subsection{Text-based Classifiers from Language Descriptions}\label{ssec:nl_temp}
\vspace{-0.1cm}
Existing OVOD approaches, \eg~Detic~\cite{Zhou2022} and ViLD~\cite{Gu2021}, make
use of simple text-based classifiers by encoding the category name with a manual prompt,
\eg~\texttt{a photo of a(n) \{class name\}} or \texttt{a(n) \{class name\}},
using an appropriate encoder --- \eg a CLIP text encoder, thereby yielding a set
of classifiers for $\mathset{C}{test}$.
This method relies on the text encoder to produce a text-based classifier
entirely from its internal understanding of \texttt{class name}.

Instead, 
we make use of natural language descriptions of categories
sourced from a large language model (LLM).
Such a design choice gives additional details like visual attributes,
leading to increased discriminative information in the classifier.
This alleviates lexical confusion --- \texttt{class name} may have two 
different meanings, and effectively prevents the need for human efforts to
manually write descriptions or spend time searching external sources for them.

Figure~\ref{fig:nl_emb} outlines our method for generating informative
text-based classifiers.
Specifically, we start by prompting an autoregressive language model with a question: 
``\texttt{What does a(n) \{class name\} look like?}'',
and sample multiple descriptions per class.
We use OpenAI's API for GPT-3~\cite{Brown2020} and generate $10$ descriptions per class with
temperature sampling~(Figure~\ref{fig:nl_emb} shows $3$ descriptions per class for clarity,
more examples can be found in Section~\ref{sec:add_cls_desc} of the Appendix),
yielding multiple descriptions of the format \texttt{\{class name\} is a \dots} or similar.
Given a set of $M$ plain text descriptions $\left\{s_i^c\right\}_{i=1}^{M}$
for class $c$,
we encode each element of the set with a CLIP text encoder~\cite{Radford2021}, 
$f_{\textsc{clip-t}}(\cdot)$, and the text-based
classifier for class $c$ is obtained from the mean of these text encodings:
\begin{align}
\mathbf{w}_{\textsc{text}}^c &= \frac{1}{M}\sum_{i=1}^{M}
f_{\textsc{clip-t}} \left(s_i^c\right)
\end{align}

At detector training time, text-based classifiers for categories of
interest~($c\in\mathset{C}{det}$ and $c\in\mathset{C}{img}$) are pre-computed and kept frozen,
the rest of detector parameters are updated \ie~all parameters
in Equations~\ref{eq:a}-\ref{eq:b}, except $\Phi_\textsc{cls}$.
At inference time, classifiers for testing categories are computed similarly
to enable open-vocabulary object detection.

\parb{Discussion.}
In this paper, 
we only consider a single straightforward question as a prompt to the LLM:
``\texttt{What does a(n) \{class name\} look like?}''.
However, it is also feasible to use alternative question prompts,  
\eg~``\texttt{How can you identify a(n) \{class name\}?}'' or
``\texttt{Describe what a(n) \{class name\} looks like?}'',
and obtain visual descriptions with the same or similar concepts~\cite{Pratt2022}.

We investigated using a transformer architecture to aggregate text embeddings
from natural language descriptions,
but we found this was not beneficial to OVOD performance over simply using the mean vector.
In general, each text embedding of a natural language description summarises
the category of interest very well and so the contrastive task,
which is used to train the aggregator (explained in detail for the visual case below),
is very easy with text embeddings yielding no improvement in text-based classifiers for OVOD.

\vspace{-0.1cm}
\subsection{Vision-based Classifiers from Image Exemplars}\label{ssec:img_temp}
\vspace{-0.1cm}
In addition to constructing classifiers with natural language descriptions,
another natural option is to use image exemplars,
especially in cases where a good description of the category is prohibitively long (such as the painted lady butterfly or \textit{Vanessa cardui} which has an intricate wing pattern),
or occasionally the class name is not known beforehand,
\eg~``Deerstalker cap'' refers to the hat often worn by Sherlock Holmes.


In such scenarios, we propose to construct classifiers by using image exemplars, 
as shown in Figure~\ref{fig:img_emb}. 
Specifically, given a set of $K$ RGB image exemplars for category $c$,
$\left\{\mathbf{x}_i^c\right\}_{i=1}^{K}$,
we encode each exemplar with a CLIP visual encoder, $f_{\textsc{CLIP-im}}(\cdot)$, 
yielding $K$ image embeddings,
which are then passed to a multi-layer Transformer~\cite{Vaswani17} with learnable [CLS] token, $\mathbf{t}_{\textsc{CLS}}$:
\begin{align}
\mathbf{w}_{\textsc{img}}^c &= \text{Transformer}\left(
\left\{f_{\textsc{CLIP-im}}\left(\mathbf{x}_i^c\right)\right\}_{i=1}^{K};
\mathbf{t}_{\textsc{CLS}}\right)
\end{align}

The Transformer architecture acts to best aggregate the $K$ image exemplars,
and the output from the [CLS] token is used as the vision-based classifier for OVOD.
When training the transformer aggregator, all exemplars are sourced from ImageNet-21k-P~\cite{Ridnik2021}.
When generating vision-based classifiers for OVOD, where possible, we source our exemplars
from ImageNet-21k~\cite{Deng09} --- where this is not possible, we use LVIS and/or VisualGenome
training data to source image exemplars.
Additional details on how we collate exemplars for training and testing are provided
in the Appendix (Section~\ref{sec:img_exem_source}).
This transformer architecture will be referred to as the \emph{visual aggregator}
and its training procedure is described next.



\parb{Offline Training. }
The visual aggregator is trained offline \ie~it is not updated during detector
training.
The training procedure needs to learn an aggregator which combines multiple
image exemplars to produce effective vision-based classifiers for OVOD
--- a classifier for a given class needs to be discriminative \wrt~other classes.
A CLIP image encoder is used to provide initial embeddings for each exemplar.
We keep the CLIP image encoder frozen during training to improve training
efficiency and prevent catastrophic forgetting in the CLIP representation.
To provide discriminative vision-based classifiers,
contrastive learning is utilised.
For a given class, the output embedding from the visual
aggregator is trained to minimise similarity with the output embedding from other classes and maximise
the similarity with an output embedding from the same class.
To do this, the contrastive InfoNCE~\cite{Oord18} loss is used.
The visual aggregator should generalise well and not be trained for a specific
downstream OVOD vocabulary,
therefore it is trained with the ImageNet-21k-P dataset~\cite{Ridnik2021} for image classification,
which contains $\sim$11M images across $\sim$11K classes.
For category $c$ during visual aggregator training, at each iteration,
two distinct sets of $K$ exemplars are sampled, augmented and encoded
by the frozen CLIP image encoder.
The two sets are input separately to visual aggregator, outputting
$2$ embeddings from the learnable [CLS] token for class $c$.
Given a batch size $B$, the InfoNCE contrastive loss ensures
sets formed from the same class have similar embeddings and those of different classes
are separated.
Once trained, the visual aggregator and visual encoder are frozen and
provide vision-based classifiers for categories in 
$\mathset{C}{det} \cup \mathset{C}{img}$/$\mathset{C}{test}$ during detector training/testing.
Additional details on how the visual aggregator is trained are provided
in the Appendix (Section~\ref{sec:img_impl_appen}).



\parb{Discussion.}
Using image exemplars for open-vocabulary detection may share some similarity to
few-shot object detection, however, there is a key distinction.
In few-shot object detection, the given ``novel/rare'' annotations (albeit few) are available for training,
\eg~recent works have found that finetuning a pre-trained object detector on few-shot detection data yields
the best results~\cite{Wang2020,Qiao2021a,Kaul22},
while in open-vocabulary detection, there are no bounding box annotations
for ``novel/rare'' categories.
Image exemplars~(\ie~the wholes image without bounding boxes) are used to specify the categories of interest;
we do not update any parameters based on
``novel/rare'' category bounding box data,
unlike in few-shot object detection.

During ablation experiments in Section~\ref{sec:ablation},
we compare the visual aggregator to a simple mean operator,
when obtaining a vision-based classifier from multiple image exemplar embeddings,
and show the benefit of the trained aggregator in this case.

\vspace{-0.1cm}
\subsection{Constructing Classifiers via Multi-Modal Fusion}\label{ssec:nl_img_temp}
\vspace{-0.1cm}
To go one step further,
a natural extension to the aforementioned methods is to construct classifiers
from multi-modal cues;
intuitively, natural language descriptions and image exemplars may contain complementary information.
For a given class, $c$,
with text-based classifier, $\classifier{text}$, 
and vision-based classifier, $\classifier{img}$, 
the multi-modal classifier, $\classifier{mm}$, 
is computed by a simple fusion method based on addition:
\begin{align}
\classifier{mm} = \frac{\classifier{text}}{\lVert\classifier{text}\rVert_{2}}
 + \frac{\classifier{img}}{\lVert\classifier{img}\rVert_{2}}
\end{align}

Figure~\ref{fig:teaser} demonstrates our entire pipeline for generating
text-based, vision-based and multi-modal classifiers, showing how any of
the three classifiers can be used with an open-vocabulary detector to detect a
``falcon''.

\parb{Discussion.}
Section~\ref{ssec:img_temp} provides details of the visual aggregator,
which yields our vision-based classifiers, but for multi-modal classifiers
we simply compute the vector sum of our $l^2$-normalised text-based and vision-based classifiers. We investigated using a unified multi-modal aggregator which ingests both text and visual embeddings, sourced from class descriptions and image exemplars, respectively. Such a model did not generate good multi-modal classifiers for OVOD
--- distinguishing between sets of text and image embeddings for different classes
becomes trivial as the text embeddings alone are sufficient to solve
the contrastive learning task, thereby ignoring the visual embeddings
altogether. Attempts to modify the training for a unified multi-modal aggregator by using Dropout~\cite{Srivastava14} on the text embeddings were not fruitful.

\vspace{-0.1cm}
\section{Experiments}\label{sec:exp}
In this section, 
we first introduce the standard dataset and benchmark used
in the literature~\cite{Gu2021,Zhou2022,Feng2022}.
Section~\ref{ssec:agg_impl} provides implementation and training details for our OVOD models,
which use classifiers constructed from natural language descriptions, visual exemplars or the combination of both.
We compare our models with existing works in Section~\ref{ssec:results},
demonstrating state-of-the-art performance.
Additionally, Section~\ref{ssec:results} provides results for cross-dataset transfer.
Section~\ref{ssec:abla_main} provides an ablation study regarding our design choices.

\vspace{-0.1cm}
\subsection{Datasets and Evaluation Protocol}\label{ssec:ovd_bench}
\parb{Standard LVIS Benchmark.} 
In this work, most experiments are based on the LVIS object detection dataset~\cite{Gupta2019},
containing a large vocabulary and a long-tailed distribution of object instances.
Specifically, the LVIS dataset contains class, bounding box and mask annotations for
1203 classes across 100k images in the MS-COCO dataset~\cite{Lin2014}.
Annotations are collected in a federated manner,
{\em i.e.}~manual annotations for a given image are not necessarily exhaustive.
The classes are divided into three sets --- rare, common and frequent
--- based on the number of training images containing a given class.

\parb{Training Datasets.}
To develop open-vocabulary object detectors, 
we follow the same setting as proposed in ViLD~\cite{Gu2021} and used in Detic~\cite{Zhou2022}.
Specifically, the original LVIS training set (\emph{LVIS-all}) is slightly changed by removing all annotations belonging to the ``rare'' categories.
This removes the annotations of 317 rare classes \emph{but not} the associated images,
in other words, objects belonging to rare categories appear in the training set but are unannotated.
This subset of LVIS training data containing only ``common'' and ``frequent''
annotations is referred to as \emph{LVIS-base}.
LVIS-base serves as $\mathset{D}{det}$ using notation from Section~\ref{ssec:prelim},
unless stated otherwise.
When using image classification data, $\mathset{D}{img}$, as extra weak supervision,
we use the subset of categories in ImageNet-21K~\cite{Deng09} that overlap with
the LVIS vocabulary and denote this subset as \emph{IN-L}, as in Detic.
IN-L covers $997$ of the $1203$ classes in the LVIS vocabulary.

\vspace{-0.1cm}
\parb{Evaluation Protocol. }
For evaluation, previous work evaluates OVOD models on the LVIS validation set (\emph{LVIS-val}) for all categories
--- treating ``rare'' classes as novel categories
as it is guaranteed that no groundtruth box annotations whatsoever are provided at the training stage.
The main evaluation metric is the standard mask AP metric averaged over the ``rare'' classes and
is denoted as APr.
The mask AP averaged across \emph{all} classes is also reported,
indicating overall class performance and is denoted as mAP.
The latter metric is an important consideration as a good model should
improve both APr and mAP; a model should not improve APr at the cost of
worse performance in terms of mAP.


\vspace{-0.1cm}
\subsection{Implementation Details}\label{ssec:agg_impl}
\parb{Object Detector Architecture.} 
The architecture we use is almost identical to that in Detic,
using the CenterNet2~\cite{Zhou2021} model with a ResNet-50 backbone~\cite{He2016}
pre-trained on ImageNet-21k-P~\cite{Ridnik2021}.
In addition to exploring different ways for constructing classifiers~($\Phi_\textsc{cls}$),
as described in Section~\ref{sec:meth},
we also add a learnable bias before generating
final confidence scores,
the effect of this bias term is investigated in the Appendix (Section~\ref{sec:ablation}).

\parb{Detector Training.}
The training recipe is the same as Detic for fair comparison,
using Federated Loss~\cite{Zhou2021} and repeat factor sampling~\cite{Gupta2019}.
While training our OVOD model on detection data only, $\mathset{D}{det}$, 
we use a $4\times$ schedule ($\sim$58 \emph{LVIS-base} epochs or 90k iterations with batch size of 64).
When using additional image-labelled data (IN-L), 
we train jointly on $\mathset{D}{det} \cup \mathset{D}{img}$ using a $4\times$ schedule
(90k iterations) with a sampling ratio of $1:4$ and batch sizes of $64$ and
$256$, respectively.
This results in $\sim$15 \emph{IN-L} epochs and an additional
$\sim$11 \emph{LVIS-base} epochs.
For mini-batches containing images from $\mathset{D}{det}$ and
$\mathset{D}{img}$ we use input resolutions of $640^2$ and $320^2$, respectively.
We conduct our experiments on $4$ 32GB V100 GPUs.

For image-labelled data, $\mathset{D}{img}$, 
an image with class label is given,
but no groundtruth bounding box is available.
Following Detic, the largest class-agnostic box proposal is used to produce an RoI feature for the given image, enabling detector training.
See the Detic paper for more details.

\parb{Text-based Classifier Construction.}
To generate plain text class descriptions we use the GPT-3 DaVinci-002 model
available from OpenAI.
For each class in LVIS,
we generate $10$ descriptions and compute the classifier
with the text encoder from a CLIP ViT-B/32 model~\cite{Radford2021},
as detailed in Section~\ref{ssec:nl_temp}.
We follow the standard method from CLIP
and use the output embedding corresponding to the final token in the input text.

\parb{Vision-based Classifier Construction.}
The visual aggregator, detailed in Section~\ref{ssec:img_temp}, 
should be general and not specific to any class vocabulary.
To fulfil this goal, 
we use the curated ImageNet-21-P dataset~\cite{Ridnik2021} as training data \emph{for the aggregator}.
This dataset, designed for pre-training visual backbones,
filters out classes with few examples from the original ImageNet-21k~\cite{Deng09} dataset,
leaving $\sim$11M images across $\sim$11K classes.

To generate a visual embedding from a given image exemplar,
we use a CLIP ViT-B/32 visual encoder.
For the aggregator, we use $N=4$ transformer blocks, with dimension $512$
(the same as the output dimension of the CLIP visual encoder) and a multilayer
perceptron dimension of $2048$.
Comprehensive details of aggregator training is provided in the
Appendix (Section~\ref{sec:img_impl_appen}).
To test the effectiveness of our visual aggregator,
we generate baseline vision-based classifiers by taking the vector mean of the
CLIP visual embeddings from the $K$ image exemplars.

When constructing vision-based classifiers for OVOD,
we find making use of test-time augmentation (TTA) improves performance.
Note, TTA here refers to augmentation of the image exemplars used to build
vision-based classifiers \emph{not} test-time augmentation of the test image on
which OVOD is performed.
In our work, each image exemplar is augmented $5$ times and input separately
to the visual encoder.
Therefore, given $K$ image exemplars, we generate $5K$ visual embeddings to be
ingested by our visual aggregator.
More details on the use of TTA in constructing vision-based classifiers are found
in the Appendix (Section~\ref{sec:ablation}).

For total clarity regarding when datasets are used
--- the visual aggregator is \emph{trained} using ImageNet-21k-P~\cite{Ridnik2021}
and the vision-based classifiers for OVOD are generated from relevant image exemplars
\emph{using the trained aggregator} (see
Section~\ref{sec:img_exem_source} of the Appendix for details on sourcing the
image exemplars).
Detection data (\eg~\emph{LVIS-base}) is only used to train the
open-vocabulary object detector and image-level data (\eg~\emph{IN-L}) may be used as
an extra source of weak supervision as in Detic~\cite{Zhou2022}.

\parb{Multi-Modal Classifier Construction.}
When computing multi-modal classifiers,
we simply compute the category-wise $l^2$-normalised classifier
from each modality,
regardless of method used to compute them, and take the vector sum.
In all cases the text-based classifiers are sourced from class descriptions
as described in Section~\ref{ssec:nl_temp}.
We combine our text-based classifiers with our vision-based classifiers,
using the trained visual aggregator.
However, once again, to test the effectiveness of our visual aggregator,
we also combine our text-based classifiers with the baseline vision-based
classifiers described in the previous paragraph.

\subsection{Open-Vocabulary Detection Results}\label{ssec:results}

\begin{table}[t]
\scriptsize
\begin{adjustbox}{width=\linewidth}
\setlength{\tabcolsep}{4pt}
\begin{tabular}{@{}l|c|c|cc@{}}
\toprule
\setlength\tabcolsep{-2pt}
\multirow{2}{*}{Model}                                  & \multirow{2}{*}{Backbone}        & Extra                    & \multirow{2}{*}{APr}        & \multirow{2}{*}{mAP}        \\
                                                        &                                  & Data                     &                             &                             \\ \midrule
ViLD~\cite{Gu2021}                                      & ResNet-50                        &                          & 16.1                        & 22.5                        \\
Detic~\cite{Zhou2022}                                   & ResNet-50                        &                          & 16.3                        & 30.0                        \\ 
ViLD-ens~\cite{Gu2021}                                  & ResNet-50                        &                          & 16.6                        & 25.5                        \\
OV-DETR~\cite{Zang2022}                                 & ResNet-50 + DETR                 &                          & 17.4                        & 26.6                        \\
F-VLM~\cite{Kuo2022}                                    & ResNet-50                        & \multirow{-5}{*}{\xmark} & \sbest{18.6}                & 24.2                        \\ \midrule
Ours (Text-Based)                                       &                                  &                          & \best{19.3}                 & \sbest{30.3}                \\
Ours (Vision-Based)                                     &                                  &                          & 18.3                        & 29.2                        \\
Ours (Multi-Modal)                                      & \multirow{-3}{*}{ResNet-50}      & \multirow{-3}{*}{\xmark} & \best{19.3}                 & \best{30.6}                 \\ \midrule \midrule
RegCLIP~\cite{Zhong2022}                                & ResNet-50                        & CC3M                     & 17.1                        & 28.2                        \\
OWL-ViT~\cite{Minderer2022}$\dagger$                    & ViT-B/32                         & LiT                      & 19.7                        & 23.3                        \\
Detic~\cite{Zhou2022}                                   & ResNet-50                        & IN-L                     & 24.6                        & 32.4                        \\ \midrule
Ours (Text-Based)                                       &                                  &                          & \sbest{25.8}                & \sbest{32.7}                \\
Ours (Vision-Based)                                     &                                  &                          & 23.8                        & 31.3                        \\
Ours (Multi-Modal)                                      & \multirow{-3}{*}{ResNet-50}      & \multirow{-3}{*}{IN-L}   & \best{27.3}                 & \best{33.1}                 \\ \midrule
\rowcolor[HTML]{EFEFEF} 
{\color[HTML]{656565} Fully-Supervised \cite{Zhou2022}} & {\color[HTML]{656565} ResNet-50} & {\color[HTML]{656565} \xmark}                 & {\color[HTML]{656565} 25.5} & {\color[HTML]{656565} 31.1} \\ \bottomrule
\end{tabular}
\end{adjustbox}
\vspace{-10pt}

\captionof{table}{
Detection performance on the LVIS Open Vocabulary Detection Benchmark using our
three types of classifier compared with previous works.
Best and second-best performing
models are coloured \best{blue} and \sbest{red}, respectively.
We split models into those which only use LVIS-base as training data (top)
and those which use additional image-level data (bottom).
Furthermore, we show results for a fully-supervised model from Detic trained on
LVIS-all in \colorbox[HTML]{C0C0C0}{\color[HTML]{656565} grey}.
$\dagger$ OWL-ViT reports bbox AP metrics and was trained on
Objects365~\cite{Shao2019} and VisualGenome~\cite{Krishna2017} {\em not} LVIS-base,
therefore it is possible LVIS-defined ``rare'' classes are contained in the
detection training data of OWL-ViT.
Due to limited compute resources we present and compare to models which
use ResNet-50~\cite{He2016} backbones or similar.
We report mask AP metrics except for $\dagger$.
\label{tab:lvis}}
\vspace{-0.1cm}
\end{table}

\parb{LVIS OVOD Benchmark.}
Table~\ref{tab:lvis} shows results on \emph{LVIS-val} for our work, 
which uses text-based, vision-based and multi-modal classifiers,
compared to a range of prior work.
We report overall mask AP performance and mask AP for ``rare'' classes only.
The latter metric is the key measure of OVOD performance.
We separate the comparisons into those models which do not use additional
image-level data (top half of Table~\ref{tab:lvis})
and those which do (bottom half of Table~\ref{tab:lvis}).
For a fair evaluation, we compare to models from prior works which use a
ResNet-50~\cite{He2016} backbone.
There are two exceptions:
(1) OWL-ViT~\cite{Minderer2022} which only
investigates using Vision Transformers for OVOD~\cite{Dosovitskiy2020} --
we compare ResNet-50 models to the ViT-B/32 OWL-ViT model as it requires similar compute during
inference in terms of GLOPs (141.5 and 139.6, respectively);
(2) OV-DETR~\cite{Zang2022} which uses a DETR-style architecture~\cite{Carion2020}
consisting of a ResNet-50 CNN backbone and modified transformer encoder and decoder.

In the experiments without using extra data~($\mathset{D}{img}=\varnothing$),
our models with text-based or multi-modal classifiers obtain
the best performance on both APr and overall mAP,
while F-VLM and Detic only performs strongly on APr and mAP, respectively.
Our model with text-based classifiers is most directly comparable to Detic and
our model outperforms Detic by $3.0$ APr.
When using extra data, we make use of a ImageNet-21k subset as in Detic (IN-L).
Models with text-based and multi-modal classifiers outperform
Detic~(previous state-of-the-art) by $1.2$ and $2.7$ APr respectively.
\textbf{Note that}, our models trained with IN-L even outperform
the fully-supervised baseline using CenterNet2,
\ie~trained on ``rare'' class box annotations.
To the best of our knowledge,
this is the first work on open-vocabulary detection that outperforms a
comparable fully-supervised model on the challenging LVIS benchmark.
Note, some other works use larger vision backbones,
\eg~Swin-B~\cite{Liu2021}, but due to limited computation resources we only
present and compare to models with ResNet-50 backbones or similar.
Our models with text-based and multi-modal classifiers surpass state-of-the-art
performance when using a ResNet-50 backbone or similar.

\begin{table}[t]
\scriptsize
\begin{adjustbox}{width=\linewidth}
\setlength{\tabcolsep}{4pt}
\begin{tabular}{@{}l|c|ccc@{}}
\toprule
\multirow{2}{*}{Model} & Extra                   & Objects365  & Objects365   & Objects365   \\
                       & Data                    & mAP         & AP50         & APr          \\ \midrule
Detic~\cite{Zhou2022}  & \multirow{2}{*}{\xmark} & 13.9        & 19.7         & 9.5          \\
Ours (Text-Based)      &                         & \best{14.8} & \best{21.0}  & \best{10.1}  \\ \midrule
Detic~\cite{Zhou2022}  & \multirow{2}{*}{IN-L}   & 15.6        & 22.2         & 12.4         \\
Ours (Text-Based)      &                         & \best{16.6} & \best{23.1}  & \best{13.1}  \\ \bottomrule
\end{tabular}
\end{adjustbox}
\vspace{-10pt}
\captionof{table}{
Detection performance when training on \emph{LVIS-all} (\ie~all LVIS training data)
and evaluating on Objects365~\cite{Shao2019}, where the least frequent $\frac{1}{3}$
of classes are defined as ``rare''.
Best performing
models are coloured \best{blue}.
Our text-based classifiers outperform Detic when transferring to Objects365 across
all classes and ``rare'' classes only.
Note the Detic~\cite{Zhou2022} models we compare to on Objects365 are not the same as the models listed in the Detic paper,
which use a large Swin-B backbone and all of ImageNet-21k as extra data for weak supervision.
We use a fair comparison Detic model which uses the same training data (see text for more details).
We report box AP metrics for Objects365.
\label{tab:o365}}
\vspace{-25pt}
\end{table}

\parb{Cross-dataset Transfer.}
Table~\ref{tab:o365} shows results for cross-dataset transfer from LVIS to Objects365~\cite{Shao2019}
when using our text-based classifiers.
We compare our work to equivalent models from Detic, reporting box AP metrics as standard in Objects365.
In all cases, these models are trained on \emph{LVIS-all} and the models in the
bottom two rows use \emph{IN-L} as extra weak supervision.
The trained open-vocabulary detectors are evaluated on the Objects365 validation set.
Following Detic, we define ``rare'' classes in Objects365 as the $\frac{1}{3}$ of classes
with the lowest frequency in the Objects365 training set.
Note the Detic models we compare to in Table~\ref{tab:o365} are not the same as those listed in Table 4 of the Detic paper,
which use a large Swin-B backbone and all of ImageNet-21k as extra data.
Instead, we compare to other Detic models publicly available which are trained on \emph{LVIS-all} (and \emph{IN-L}).
For ease, we provide links to the Detic configuration and checkpoint files used in this cross-dataset transfer evaluation
\footnote{Detic
    \href{https://github.com/facebookresearch/Detic/blob/main/configs/BoxSup-C2_L_CLIP_R5021k_640b64_4x.yaml}{Configuration} 
    \href{https://dl.fbaipublicfiles.com/detic/BoxSup-C2_L_CLIP_R5021k_640b64_4x.pth}{Checkpoint} Extra Data: \xmark}
\footnote{Detic
    \href{https://github.com/facebookresearch/Detic/blob/main/configs/Detic_LI_CLIP_R5021k_640b64_4x_ft4x_max-size.yaml}{Configuration} 
    \href{https://dl.fbaipublicfiles.com/detic/Detic_LI_CLIP_R5021k_640b64_4x_ft4x_max-size.pth}{Checkpoint} Extra Data: \emph{IN-L}}.
Evaluation on Objects365 after training on LVIS is easily done with Detic or with our text-based classifiers.
In the case of Detic, the simple classifiers based on LVIS class names are replaced with equivalent simple classifiers based
on Objects365 class names.
For our text-based classifiers, plain text descriptions are generated for each of the Objects365 classes and
are encoded as described in Section~\ref{ssec:nl_temp} and replace the text-based classifiers for LVIS.
In both models, Detic and ours, all other parameters of the open-vocabulary detector remain the same.
In all cases, using our text-based classifiers gives performance improvements over the equivalent Detic model.
Considering all classes, our method with extra data outperforms Detic by $1.0$ mAP and $0.9$ AP50.
For ``rare'' classes as defined above, our method with extra data outperforms the equivalent Detic model by $0.7$ APr.
These results demonstrate that our method, which uses text-based classifiers generated from rich class descriptions,
provides additional information compared to using a simple classifier based on the class name only,
even when the training and testing class vocabularies are disjoint.

\vspace{-0.1cm}
\subsection{Ablation Study}\label{ssec:abla_main}
\vspace{-0.1cm}
\begin{figure*}[t]
\centering
\scriptsize
\includegraphics[width=1.0\linewidth]{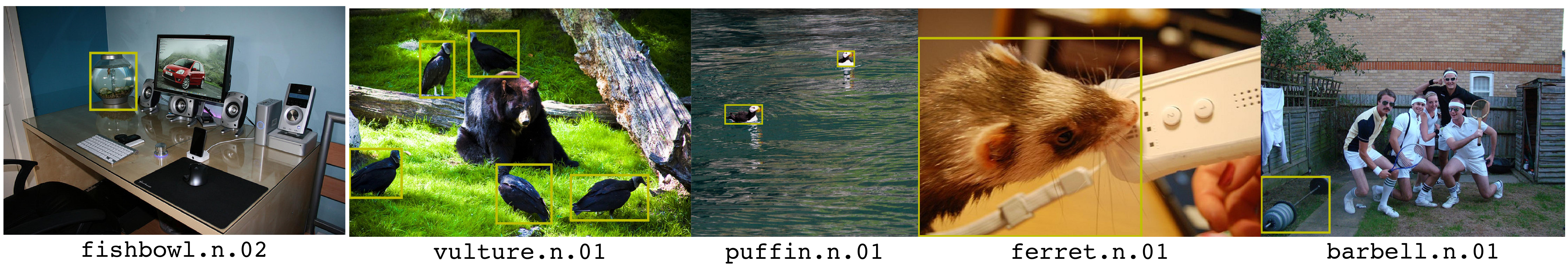}
\label{fig:qual_eg}
\vspace{-0.6cm}
\caption{Some qualitative detection examples using our model with text-based classifiers, detecting ``rare'' category instances in \emph{LVIS-val}.
Our text-based classifiers are sourced from rich natural language descriptions of a given class by prompting an GPT-3 LLM.}
\vspace{-10pt}
\end{figure*}

\parb{Results with Vision-Based Classifiers.}
Using vision-based classifiers for OVOD is an under-explored area and
so we compare our method,
detailed in Section~\ref{ssec:img_temp},
to baseline classifiers in which the same visual encoder is used,
but the action of the aggregator is replaced by performing simple vector mean.
The \colorbox[HTML]{FFCE93}{orange} rows of Table~\ref{tab:lvis_cons} compares
the use of our visual aggregator to performing simple vector mean across visual
embeddings instead.
When no additional image-level data is
used~(top two \colorbox[HTML]{FFCE93}{orange} rows) our aggregator (Model B)
boosts performance by $3.5$ APr compared to the vector mean baseline (Model A).
For models which train on additional image-level data (IN-L) our aggregator (Model G) boosts
performance by $2.2$ APr compared to the baseline (Model F).
This comparison demonstrates the utility of our
visual aggregator in constructing better vision-based classifiers
rather than na\"ively averaging the $K$ visual embeddings.
Note our vision-based classifiers and the baseline classifiers both utilise TTA
as mentioned in Section~\ref{ssec:agg_impl} (see Section~\ref{sec:ablation} of
the Appendix for details on TTA).
The results in Table~\ref{tab:lvis_cons} use $K=5$.
Further results for $K=1,2,10$ are found in the Appendix
(Section~\ref{sec:multi_k}).


\begin{table}[t]
\begin{adjustbox}{width=0.96\linewidth, center}
\setlength{\tabcolsep}{12pt}
\begin{tabular}{c|c|c|c|c|cc@{}}
\toprule
               & Visual     & Visual     & Text       & Extra                               &              &              \\
Model          & Mean?      & Agg.?      & Cls.?      & Data?                               & APr          & mAP          \\ \midrule
\rowcolor[HTML]{FFCE93} 
A & \checkmark &            &            & \cellcolor[HTML]{FFCE93}                         & 14.8         & 28.8         \\
\rowcolor[HTML]{FFCE93} 
B &            & \checkmark &            & \multirow{-2}{*}{\cellcolor[HTML]{FFCE93}\xmark} & 18.3         & 29.2         \\ \midrule
\rowcolor[HTML]{DAE8FC} 
C &            &            & \checkmark & \xmark                                           & \sbest{19.3} & 30.3         \\ \midrule
\rowcolor[HTML]{C0C0C0}
D & \checkmark &            & \checkmark &                                                  & \best{20.7}  & \sbest{30.5} \\
\rowcolor[HTML]{C0C0C0}
E &            & \checkmark & \checkmark & \multirow{-2}{*}{\xmark}                         & \sbest{19.3} & \best{30.6}  \\ \midrule \midrule
\rowcolor[HTML]{FFCE93} 
F & \checkmark &            &            & \cellcolor[HTML]{FFCE93}                         & 21.6         & 31.3         \\
\rowcolor[HTML]{FFCE93} 
G &            & \checkmark &            & \multirow{-2}{*}{\cellcolor[HTML]{FFCE93}IN-L}   & 23.8         & 31.3         \\ \midrule
\rowcolor[HTML]{DAE8FC} 
H &            &            & \checkmark & IN-L                                             & 25.8         & 32.7         \\ \midrule
\rowcolor[HTML]{C0C0C0}
I & \checkmark &            & \checkmark &                                                  & \sbest{26.5} & \sbest{32.8} \\
\rowcolor[HTML]{C0C0C0}
J &            & \checkmark & \checkmark & \multirow{-2}{*}{IN-L}                           & \best{27.3}  & \best{33.1}  \\ \bottomrule
\end{tabular}
\end{adjustbox}

\scriptsize
\captionof{table}{
Detection performance on the LVIS OVOD benchmark comparing all three of our methods:
(1) \colorbox[HTML]{FFCE93}{orange} --- vision-based classifiers;
(2) \colorbox[HTML]{DAE8FC}{blue} --- text-based classifiers;
(3) \colorbox[HTML]{C0C0C0}{grey} --- multi-modal classifiers.
Results for models trained only on LVIS-base and LVIS-base+IN-L
are shown in the top and bottom halves, respectively.
Visual Mean?: \emph{simple vector mean} is used to combine
visual embeddings of image exemplars,
Visual Agg.?: \emph{our visual aggregator} is used to combine
visual embeddings,
Text Cls.?: text-based classifiers are used.
Models which use text-based and vision-based classifiers represent our models
with multi-modal classifiers.
We report mask AP metrics.
}\label{tab:lvis_cons}
\vspace{-8pt}
\end{table}

\parb{Results with Multi-Modal Classifiers.}
To evaluate the effectiveness of multi-modal classifiers
we perform similar experiments
as those using vision-based classifiers,
except for each model we combine the vision-based classifiers with the
text-based classifiers,
as described in Section~\ref{ssec:nl_img_temp}.
Results for multi-modal classifiers are shown in \colorbox[HTML]{C0C0C0}{grey}
rows of Table~\ref{tab:lvis_cons}.
With no additional image-level data,
the vector mean baseline (Model D)
outperforms the use of our aggregator (Model E) by $1.4$ APr.
However, for models with image-level data (IN-L) our aggregator (Model J) boosts
performance by $0.8$ APr compared to the baseline (Model I).
Furthermore, comparing the multi-modal classifiers
(\colorbox[HTML]{C0C0C0}{grey} rows in Table~\ref{tab:lvis_cons})
with text-based classifiers (\colorbox[HTML]{DAE8FC}{blue} rows in Table~\ref{tab:lvis_cons})
demonstrates that in all cases adding information from image exemplars yields
improved OVOD performance --- our best multi-modal model improves performance
over our best text-based model by $1.5$ APr confirming that combining
the vision and text modalities utilises complementary information between
the two.

\parb{Relationship between IN-L and LVIS ``rare'' classes.}
Section~\ref{ssec:lvis_val_plus} of the Appendix splits the APr metric into two
based on the ``rare'' LVIS categories contained in IN-L.
One may expect improvements in APr performance when training on
IN-L to only come from ``rare'' categories found in IN-L.
Our evaluation finds this not to be the case.
Detailed results can be found in the Appendix (Section~\ref{ssec:lvis_val_plus})
which breaks the APr metric into ``rare'' categories found in IN-L and those not.

\parb{Additional Ablation Experiments.}
Section~\ref{sec:ablation} of the Appendix presents ablation experiments which demonstrate:
(1) applying a learnable bias before calculating the final detection score for a region
improves OVOD performance;
(2) improvements in OVOD performance using our text-based classifiers is orthogonal to
applying this learnable bias;
(3) applying TTA on image exemplars yield better vision-based classifiers for OVOD;
(4) comparisons between our text-based classifiers and those generated from manual prompts.
Please refer to Section~\ref{sec:ablation} of the Appendix for details and evaluation results
for these experiments.

\vspace{-0.2cm}
\section{Conclusion}\label{sec:conc}
\vspace{-0.1cm}
In this paper, we tackle open-vocabulary object detection by
investigating the importance of the method used to generate classifiers for
open-vocabulary detection.
This work goes beyond the very simple methods used in prior work to generate
such classifiers --- with the class name only.
We present a novel method which combines a large language model~(LLM) and a visual-language model~(VLM) to produce improved classifiers.
Moreover, we investigate using image exemplars to provide classifiers
for OVOD and present a method for generating such classifiers using a large
classification dataset and a simple transformer based architecture.
Finally, we combine our classifiers from the two modalities to produce
multi-modal classifiers for OVOD.
Our experiments show that our method using natural language only outperforms current state-of-the-art OVOD works,
especially in cases where no extra image-level data is used.
Furthermore, our multi-modal classifiers set new state-of-the-art performance
with a large improvement over prior work.

\vspace{-0.2cm}
\section*{Acknowledgements}
\vspace{-0.1cm}
We thank Lauren Bates-Brownsword for help with proofreading.
This research was supported by the EPSRC CDT in AIMS EP/L015897/1 and EP/S515541/1,
the EPSRC Programme Grant VisualAI EP/T028572/1,
and a Royal Society Research Professorship.
Weidi Xie would like to acknowledge the National Key R\&D Program of China~(No. 2022ZD0161400).


{
    \clearpage
    \small
    \bibliographystyle{icml2023}
    \bibliography{longstrings,vgg_local,vgg_other}
}

\newpage
\appendix
\onecolumn
\section{Ablation Studies}\label{sec:ablation}

We now ablate some of the key components using the open-vocabulary LVIS
benchmark without any extra image classification data
\ie~$\mathset{D}{img} = \varnothing$, unless stated otherwise.
All metrics have the standard LVIS definition and we report mask AP metrics in all cases.
For reference APr, APc and APf represent mean average precision across ``rare'', ``common''
and ``frequent'' classes, respectively, as defined in LVIS.
Moreover, mAP, AP50 and AP75 represent mean average precision across all classes but
for all intersection-over-union (IoU) criteria, IoU$=0.5$ and IoU$=0.75$, respectively.

\begin{table}[h!]
\centering
\begin{adjustbox}{width=0.52\linewidth}
\begin{tabular}{@{}cc|ccc|ccc@{}}
\toprule
\setlength\tabcolsep{0pt}
Bias?       & Init. Value & mAP  & AP50 & AP75 & APr  & APc  & APf  \\ \midrule
\xmark      & N/A         & 29.7 & 43.7 & 31.6 & 15.6 & 30.5 & 35.0 \\
\checkmark  & -2.0        & 29.9 & 43.5 & 32.0 & 17.7 & 30.1 & 35.0 \\ \bottomrule
\end{tabular}
\end{adjustbox}
\captionof{table}{
The effect of including a learnable bias on detections scores for OVOD.
The top row does not use a learnable bias, as in Detic.
The bottom row applies a learnable bias prior to computing final detection scores
with logistic sigmoid.
Applying a learnable bias improves performance on novel/rare categories (APr).
We report mask AP metrics.
}\label{tab:bias}
\end{table}

\parb{Effect of Detection Score Bias.}
Table~\ref{tab:bias} shows the effect of adding a learnable bias to the
detection scores before applying a logistic sigmoid to get a final detection
score in the range $[0,1]$
To evaluate the effect of the learnable bias only,
our proposed text-based classifiers sourced from rich class descriptions,
as described in Section~\ref{ssec:nl_temp},
\emph{are not} used and instead the same simple
text-based classifiers used in Detic, of form ``\texttt{a(n) \{class name\}}'',
are used in this comparison.
We observe adding a learnable bias improves open-vocabulary detection by
$2.1$ AP on rare categories compared to not using a bias, as done in Detic.
Without the use of a bias, class-agnostic proposals are not biased towards being
labelled as background.
With respect to a given class,
a proposal is most likely to be negative,
therefore use of a bias makes intuitive sense to reflect this and
stabilises early training of the detector.
Similar findings were found in RetinaNet~\cite{Lin2017a}.

\begin{table}[h!]
\centering
\begin{adjustbox}{width=0.5\linewidth}
\begin{tabular}{@{}c|ccc|ccc@{}}
\toprule
Model            & mAP  & AP50 & AP75 & APr  & APc  & APf  \\ \midrule 
Detic            & 30.2 & 44.2 & 32.1 & 16.4 & 31.0 & 35.4 \\
Ours (w/o bias)  & 30.4 & 44.4 & 32.3 & 18.6 & 30.8 & 35.2 \\
Ours (w/ bias)   & 30.3 & 44.2 & 32.2 & 19.3 & 30.5 & 35.0 \\ \bottomrule       
\end{tabular}
\end{adjustbox}
\captionof{table}{
The effect of using our text-based classifiers sourced from rich descriptions.
In contrast, Detic uses simple classifiers based on class names only (top row).
Results for a detector trained using our text-based classifiers but no learnable bias is
shown in the middle row.
Our proposed model makes use of text-based classifiers sourced from rich descriptions
and a learnable bias (bottom row).
Our method for text-based classifiers
improves performance on novel/rare categories (APr) by a large amount.
We report mask AP metrics.
}\label{tab:gpt3}
\end{table}

\parb{Natural Language Descriptions.}
Table~\ref{tab:gpt3} shows the effect of using rich class descriptions,
sourced from a LLM, rather than forming text-based classifiers
from simple text prompts of the format
``\texttt{a(n) \{class name\}}'' as in Detic.
To compare fairly to Detic with detection data only (top row),
we report a set of results which do not make use of the
learnable bias on the detection scores as detailed above (middle row).
Using our text-based classifiers without a learnable bias improves performance on rare categories by
$2.2$ APr compared to the public Detic model.
Using rich class descriptions, a learnable bias and our method (bottom row) further improves
open-vocabulary detection on novel/rare categories by $2.9$ and $0.7$ APr compared to
the public Detic model and our method without a learnable bias, respectively.

\begin{table}[h!]
\centering
\begin{adjustbox}{width=0.45\linewidth}
\begin{tabular}{@{}c|c|cccc@{}}
\toprule
\setlength\tabcolsep{0pt}
Visual Encoder                        & TTA?               & mAP  & APr  & APc  & APf  \\ \midrule
\multirow{3}{*}{CLIP ViT-B/32}        & \xmark             & 29.0 & 16.3 & 29.1 & 34.4 \\
                                      & \checkmark, harsh  & 29.0 & 17.2 & 28.7 & 34.6 \\
                                      & \checkmark, gentle & 29.2 & 18.3 & 28.7 & 34.4 \\ \bottomrule
\end{tabular}
\end{adjustbox}
\captionof{table}{The effect of using test-time augmentation (TTA) when
generating classifier embeddings from image exemplars using a CLIP image encoder.
Both TTA recipes use two common augmentations --- \texttt{ColorJitter} and
\texttt{RandomHorizontalFlip}.
For harsh/gentle TTA --- min scale of \texttt{RandomResizedCrop} $= 0.5/0.8$.
Both harsh and gentle TTA perform better than no TTA in terms of performance
on novel/rare categories (APr).
We report mask AP metrics.
}\label{tab:tta}
\end{table}

\parb{Test-Time Augmentation on Image Exemplars for Vision-Based Classifiers.}
Table~\ref{tab:tta} shows the effect of using test-time augmentation (TTA)
on image exemplars to produce vision-based classifiers with our trained
aggregator.
For each image exemplar we generate $5$ augmentations.
As in the main paper, we use the case of $K=5$ --- for each class in the LVIS vocabulary
we are have $5$ RGB image exemplars.
As mentioned in Section~\ref{ssec:agg_impl},
we augment each exemplar $5$ times when using TTA.
We consider two augmentation variations, with each containing
\texttt{ColorJitter} and \texttt{RandomHorizontalFlip}.
The `harsh' variation uses \texttt{RandomResizedCrop(scale=(0.5,1.0))}
and the `gentle' variation uses \texttt{RandomResizedCrop(scale=(0.8,1.0))}.
We find adding `gentle' TTA performs best, improving open-vocabulary detection
by $2.0$ AP on rare categories compared to no use of TTA.
In the main paper, when using vision-based classifiers we utilise `gentle' TTA
on the image exemplars.

\begin{table}[h!]
\centering
\begin{adjustbox}{width=0.55\linewidth}
\begin{tabular}{@{}c|c|cccc@{}}
\toprule
Prompt                                           & mAP  & APr  & APc  & APf  \\ \midrule
a/an \texttt{class name}                         & 29.9 & 17.7 & 30.1 & 35.0 \\
a photo of a/an \texttt{class name}              & 29.5 & 15.9 & 30.0 & 34.9 \\
a photo of a/an \texttt{class name} in the scene & 29.4 & 16.2 & 29.7 & 34.8 \\ \midrule
Our LLM descriptions                             & 30.3 & 19.3 & 30.5 & 35.0 \\ \bottomrule
\end{tabular}
\end{adjustbox}
\captionof{table}{The effect of using manually crafted prompts against our rich class descriptions sourced
from LLMs. All models use a learnable bias on the detection scores and text-based classifiers.
Using our class descriptions improves performance on ``rare'' classes compared to manually crafted prompts.
}\label{tab:prompt}
\end{table}

\parb{Comparing our LLM Descriptions to Manually Designed Text Prompts.}
Table~\ref{tab:prompt} compares the detector performance between using simple manually
crafted prompts (first three rows) and our rich class descriptions sourced from an LLM (final row),
for constructing text-based classifiers.
We report results for the case where the detector is trained on \emph{LVIS-base} only (\ie~no
additional image-level data is used).
In all cases we apply the learnable bias on the detection score.
We note that of the manual prompts, the simplest one, of the form ``\texttt{a(n) \{class name\}}'',
performs best across all metrics.
Using our text-based classifiers generated from LLM descriptions improves performance on
rare classes by $1.6$ APr and by $0.4$ mAP across all classes.
Performance on ``common'' and ``frequent'' classes is largely similar as the availability of
labelled detection data for these classes renders the quality of the classifier less important.
\newpage
\section{A Closer Look at ``rare'' Class Performance}\label{ssec:lvis_val_plus}
\begin{table}[h!]
\centering
\begin{adjustbox}{width=0.45\linewidth}
\begin{tabular}{@{}c|c|cccc@{}}
\toprule
\setlength\tabcolsep{0pt}
Model                 & $\mathset{D}{img}$          & mAP           & APr           & APr-w         & APr-z         \\ \midrule
Detic~\cite{Zhou2022} & \multirow{3}{*}{\xmark}     & 30.2          & 16.3          & 15.7          & 19.7          \\
Ours (Text-Based)     &                             & 30.3          & \textbf{19.3} & \textbf{19.2} & 19.4          \\
Ours (Multi-Modal)    &                             & \textbf{30.6} & 19.2          & 18.5          & \textbf{22.2} \\ \midrule
Detic~\cite{Zhou2022} & \multirow{3}{*}{\checkmark} & 32.4          & 24.9          & 25.4          & 23.0          \\
Ours (Text-Based)     &                             & 32.6          & 25.8          & 26.7          & 21.7          \\
Ours (Multi-Modal)    &                             & \textbf{33.1} & \textbf{27.3} & \textbf{27.8} & \textbf{24.9} \\ \bottomrule
\end{tabular}
\end{adjustbox}
\captionof{table}{Comparison between Detic and our models using text-based classifiers
and multi-modal classifiers on LVIS.
As IN-L does not cover all classes in the LVIS vocabulary,
we split the rare class metric APr into APr-w and APr-z
which represent rare classes with and without weak annotations from IN-L,
respectively.
Best performing models are shown in \textbf{bold}.
We report mask AP metrics.
}\label{tab:rare_breakdown}
\end{table}

Section~\ref{ssec:ovd_bench} detailed the data used to train our detector.
IN-L contains images from the $997$ classes
from LVIS present in ImageNet-21k~\cite{Deng09}.

IN-L gives weak supervision during detector training.
Out of the $337$ ``rare'' classes in LVIS,
$277$ are covered by IN-L and are therefore weakly supervised,
leaving $60$ classes for which no weak supervision is available.

To investigate the improvement in performance when using IN-L,
we split the rare class metric (APr), which reports average precision across rare classes,
into APr-w which averages across rare classes \emph{present} in IN-L
and APr-z which averages across rare classes \emph{not present} in IN-L which are therefore
truly zero-shot classes.
Note that when no extra image-level data is used,
\ie~$\mathset{D}{img}=\varnothing$, all rare classes are truly zero-shot classes.

Table~\ref{tab:rare_breakdown} shows the result of using this breakdown.
These results show training on IN-L improves performance on rare classes
not contained in IN-L, which may not be expected.
The weak supervision from IN-L leads to a reduction in false positives for all
rare classes leading to improved performance across all metrics.
Moreover, our multi-modal classifiers perform best across all metrics.

\newpage
\section{Vision-based Classifier Pipeline Implementation Details}\label{sec:img_impl_appen}
For the transformer architecture of the visual aggregator detailed in
Section~\ref{ssec:img_temp}, we use $4$ transformer encoder blocks,
each with a hidden dimension of $512$ and MLP dimension $2048$.
As input to the first transformer block, we encode each image exemplar with
a CLIP image encoder which remains frozen throughout training,
yielding one embedding per exemplar.
The set of embeddings are input with a learnable [CLS] token. The output
[CLS] token is used as the final vision-based classifier.

To train the model we use the ImageNet-21k-P dataset~\cite{Ridnik2021} for 10 epochs.
To speed up and improve training we store a dynamic queue of
size~$4096\times{K}$ CLIP encoded embeddings, with $512$ positions in the
queue updated each iteration, using a last-in first-out policy.
Each set of $K$ represents encodings from $K$ randomly sampled images
for a single class. We use $K=5$.
For contrastive training, we use a temperature of $0.02$
in the InfoNCE~\cite{Oord18} loss function,
the AdamW~\cite{Loshchilov2018} optimiser with standard hyperparameters and a
learning rate of $0.0002$.
Furthermore, during training we uniformly sample $k\in[1:K]$ to simulate
varying numbers of image exemplars being available for
downstream OVOD when $\mathset{C}{test}$ is defined.
Therefore, for a given iteration, there may $1-5$, visual embeddings
input per class.
Prior to input to the CLIP image ViT encoder,
we apply random augmentations to each sampled image from ImageNet-21k-P.
We use an augmentation policy similar to SimCLR~\cite{Chen2020},
which includes \texttt{RandomResizedCrop}, \texttt{ColorJitter}
and \texttt{RandomGrayscale}.
We discover that test-time augmentation of the image exemplars available
for the vocabulary in $\mathset{C}{test}$ improves downstream OVOD performance.
For each image exemplar, we generate $5$ test-time augmentations.
Therefore if we have $L$ image exemplars for a given class in
$\mathset{C}{test}$, $5L$ augmented images are encoded using the CLIP image
encoder and fused using the learnt transformer architecture --- the visual
aggregator --- as described in Section~\ref{ssec:img_temp}.
Use of test-time augmentation is ablated in Section~\ref{sec:ablation}.

\newpage
\section{Sourcing Image Exemplars}\label{sec:img_exem_source}

In this section, we detail how image exemplars are sourced when performing
experiments using vision-based classifiers and multi-modal classifiers,
as described in Section~\ref{ssec:img_temp} and \ref{ssec:nl_img_temp},
respectively. We start with an empty image exemplar dictionary (IED) for the $1203$
LVIS classes.

From constructing the image-level dataset IN-L,
we know that ImageNet-21k~\cite{Deng09} contains $997$ out of the
$1203$ classes in the LVIS vocabulary,
using exact WordNet~\cite{Miller1995} synset matching.
We add IN-L to our IED.
The result is $988$ classes have more than $40$ images in ImageNet-21k.
This leaves $215$ classes for which there are too few or no image
exemplars ($<40$).

Next, to try and fill this gap, we turn to LVIS itself.
We add the LVIS training annotations with area greater
than $32^2$ to our IED.
There are now $1095$ LVIS classes with more than $40$ examples,
leaving $48$ classes with at least $10$ exemplars and $60$ classes
with less than $10$ exemplars.

The final dataset we turn to is VisualGenome~\cite{Krishna2017}, which provides
bounding box annotations for $7605$ WordNet synsets.
We include the annotations from VisualGenome with an exact WordNet synset match
with the LVIS vocabulary to our IED.
We now have $1110$ LVIS classes with at least $40$ exemplars and $1160$ with
at least $10$ exemplars.
Reducing our minimum required number of exemplars per class from $40$ to $10$
leaves $43$ classes with too few exemplars.

At this point, we inspect each of the remaining $43$ classes by hand and find
that all have other synsets present in ImageNet-21k which are visually identical
or very similar.
For example, ``anklet'' is a ``common'' class in LVIS, for which LVIS gives a
definition of ``an ornament worn around the ankle'' and a WordNet synset of
\texttt{anklet.n.03}.
This synset is not found in ImageNet-21k but \texttt{anklet.n.02},
defined as ``a sock that reaches just above the ankle'' by WordNet, is present
and visual inspection shows these images to
actually exactly match \texttt{anklet.n.03}.
Therefore, we add ImageNet-21k images relating to \texttt{anklet.n.02} to our IED.
As another example, \texttt{penny.n.02} (as in the penny coin) is a ``rare''
class in LVIS for which exemplars could not be found automatically.
However, ImageNet-21k contains images of \texttt{coin.n.01} which is a hypernym
\texttt{penny.n.02}.
The images for \texttt{coin.n.01} are visually
extremely similar and often identical to those one would expect for \texttt{penny.n.02}
and so we add ImageNet-21k images relating to \texttt{coin.n.02} to our IED.

After applying some human effort as described above,
our IED contains at least $40$ image exemplars for $1110$ (92\% of LVIS classes)
and at least $10$ image exemplars for all LVIS classes.

\newpage
\section{Varying Number of Image Exemplars for Vision-based Classifiers}\label{sec:multi_k}

\begin{figure}[t]
\centering
\scriptsize
\includegraphics[width=0.7\linewidth]{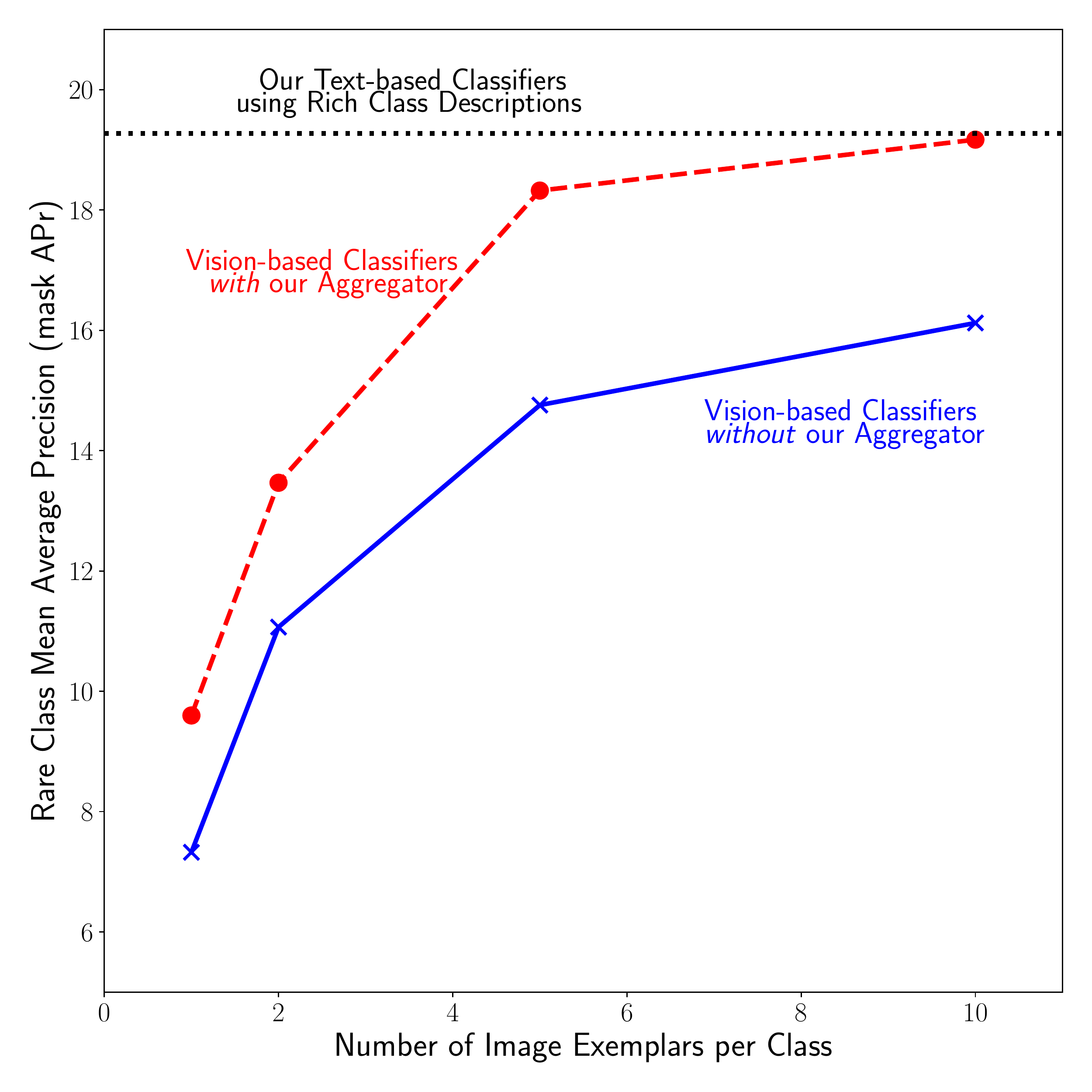}
\caption{Detection performance of our vision-based classifiers on the LVIS OVOD
benchmark. We vary the number of image exemplars available per class, $K$,
to investigate the effect of the number of image exemplars on OVOD performance.
}\label{fig:multi_k}
\end{figure}

In this section we show results using vision-based classifiers varying 
the number of $K$ image exemplars used per class.
Figure~\ref{fig:multi_k} shows performance on the LVIS OVOD benchmark for
rare classes using $K=1,2,5,10$,
where $K$ is the number of image exemplars per class used.
We compare our method which makes use of our aggregator (red dashed),
which has a transformer architecture, with the simple vector mean
of the embeddings (blue solid) for the $K$ image exemplars.
In both cases we apply the `gentle' TTA detailed and ablated in Section~\ref{sec:ablation}.

These results use LVIS-base as detection training data, no additional image-level
labelled data \ie~$\mathset{D}{img}=\varnothing$ and CLIP ViT-B/32 as the
pre-trained visual encoder to produce initial embeddings from each exemplar.

Figure~\ref{fig:multi_k} shows that for each value of $K$,
the use of our aggregator boosts performance on rare classes demonstrating
the utility of our aggregator at combining the most useful information from the
$K$ given exemplars.
Our method for $K=5$ surpasses the performance of $K=10$ with simple vector
averaging.
For $K=1$, our method improves performance by $2.3$ APr which further
demonstrates the improved feature representation --- $K=1$ involves no
aggregation as only $1$ exemplar is available per class.

Furthermore, we compare to the performance of our text-based classifiers which
make use of rich class descriptions sourced from a GPT-3 model.
Our vision-based classifiers cannot surpass the performance of our text-based
classifiers demonstrating the need for more research into using image exemplars
for OVOD.

\newpage
\section{Additional Example Class Descriptions}\label{sec:add_cls_desc}

In this section we show a selection of rich class descriptions sourced from the
\texttt{text-davinci-002} text completion model from OpenAI.
For each class in the LVIS vocabulary we generate $10$ rich descriptions.
We also give the LVIS frequency category --- rare, common, frequent.

\subsection{Generated descriptions for ``bagpipe'' (rare)}
\vspace{-25pt}
\texttt{
    {\begin{enumerate}
        \footnotesize
        \setlength{\itemsep}{0pt}%
        \setlength{\parskip}{0pt}%
        \item A bagpipe is a wind instrument with a bag that is filled with air.
        \item A bagpipe typically consists of a blowstick, a chanter, and one or more drones.
        \item A bagpipe is a musical instrument that has a bag, a blowpipe, and usually two drones.
        \item A bagpipe is a wind instrument with a bag that collects air, a reed pipe for each note, and a blowpipe.
        \item A bagpipe is a musical instrument that is played by blowing into a bag of air.
        \item Bagpipes vary in appearance, but most have a bag made from a animal skin, a blowpipe, a chanter, and one or more drones.
        \item A bagpipe is a musical instrument that has a bag, a blowpipe, and usually drone pipes.
        \item A bagpipe is a musical instrument that is usually made out of wood.
        \item A typical Highland bagpipe has a chanter with a double reed, a blowstick, three drones with single reeds, and a bag.
        \item A bagpipe consists of a blowing bag, a chanter, a drone, and usually one or more drones.
    \end{enumerate}}
}

\subsection{Generated descriptions for ``trench coat'' (rare)}
\vspace{-25pt}
\texttt{
    {\begin{enumerate}
        \footnotesize
        \setlength{\itemsep}{0pt}%
        \setlength{\parskip}{0pt}%
        \item A trench coat is a coat made of heavy cloth, sometimes waterproof, that hangs to about knee length.
        \item A trench coat looks like a long, military-style coat.
        \item A trench coat typically looks like a long, belted raincoat.
        \item A trench coat is a long, water-resistant coat that is typically worn over top of a suit.
        \item A trench coat typically has a removable liner, a double-breasted front, and belted cuffs.
        \item A trench coat generally refers to a type of coat that is longer than waist length.
        \item A trench coat is a coat that is usually a little bit longer than waist length, has a tie or a belt around the waist, and has a collar.
        \item A trench coat is a coat made of waterproof material, typically hip-length or longer, with a belt and a collar.
        \item A trench coat is a long, light coat with a belt.
        \item A trench coat is a raincoat made of heavy-duty fabric, typically poplin, gabardine, or drill.
    \end{enumerate}}
}

\subsection{Generated descriptions for ``walrus'' (rare)}
\vspace{-25pt}
\texttt{
    {\begin{enumerate}
        \footnotesize
        \setlength{\itemsep}{0pt}%
        \setlength{\parskip}{0pt}%
        \item A walrus is a large, flippered marine mammal with a bulky body, short limbs, and a large head with two long tusks protruding from the mouth.
        \item A walrus is a blubbery mammal with long tusks, whiskers, and a seal-like face.
        \item A walrus is a large, flippered marine mammal with a long, tusked head.
        \item A walrus is a stocky, rounded pinniped with small flippers, short fur, and long tusks.
        \item A walrus is a large, flippered marine mammal with a bulky body, short tail, and wide, flat head.
        \item A walrus is a large ocean mammal with two long tusks, a thick fur coat, and large flippers.
        \item A walrus is a large flippered marine mammal with a discontinuous distribution about the North Pole in the Arctic Ocean and sub-Arctic seas of the Northern Hemisphere.
        \item A walrus is a large flippered marine mammal with a thick fur coat.
        \item A walrus is a large marine mammal with a body shaped somewhat like a seal.
        \item A walrus is a seal with a long face and large tusks.
    \end{enumerate}}
}

\newpage

\subsection{Generated descriptions for ``briefcase'' (common)}
\vspace{-25pt}
\texttt{
    {\begin{enumerate}
        \footnotesize
        \setlength{\itemsep}{0pt}%
        \setlength{\parskip}{0pt}%
        \item A briefcase is a rectangular, portable case used to hold papers, documents, or other materials.
        \item A briefcase is a small case used to carry documents and other small items.
        \item A briefcase is a small, rectangular-shaped case that is used to carry important documents or other items.
        \item A briefcase is typically a rectangle shaped bag made of leather or synthetic materials.
        \item A briefcase generally looks like a small, rectangular case made out of a variety of materials, such as leather, canvas, or nylon.
        \item A briefcase is a narrow rectangular case used to carry documents and other valuables.
        \item A briefcase is a box-shaped bag typically used by businesspeople to transport important documents.
        \item A briefcase is a rectangular leather case with a handle.
        \item A briefcase is a small case used to carry documents and other small items.
        \item A typical briefcase is rectangular and has a handle on the top.
    \end{enumerate}}
}

\subsection{Generated descriptions for ``coin'' (common)}
\vspace{-25pt}
\texttt{
    {\begin{enumerate}
        \footnotesize
        \setlength{\itemsep}{0pt}%
        \setlength{\parskip}{0pt}%
        \item A coin is a small, flat, round piece of metal or plastic that is used as money.
        \item A coin has a head side and a tail side.
        \item A coin is usually a small, flat, round piece of metal or plastic that is used as money.
        \item A coin has a round shape and is flat.
        \item A coin generally has a circular shape with a raised edge, and two faces --- one on each side.
        \item A coin is a small, flat, round piece of metal or plastic that is used as money.
        \item A coin is a round piece of metal with an image on one side and the words ``United States of America'' on the other.
        \item A coin is a small, round, flat piece of metal or plastic that is used as money.
        \item Sure, a coin is a small, round, flat piece of metal or plastic that is used as money.
        \item A coin is a small, round, metal disk with an image on one side and raised lettering on the other.
    \end{enumerate}}
}

\subsection{Generated descriptions for ``waffle'' (common)}
\vspace{-25pt}
\texttt{
    {\begin{enumerate}
        \footnotesize
        \setlength{\itemsep}{0pt}%
        \setlength{\parskip}{0pt}%
        \item A waffle is a pancake-like pastry that is cooked in a waffle iron and has a distinctively grid-like pattern on the top and bottom.
        \item A waffle is a dish made from leavened batter or dough that is cooked between two plates that are patterned to give a characteristic size, shape, and surface impression.
        \item A waffle is a thin, round, batter-based cake that is cooked in a waffle iron and is usually served with syrup.
        \item A waffle looks like a round, honeycomb-patterned cake that is cooked in a waffle iron.
        \item A waffle is typically a leavened batter or dough that is cooked between two plates that are patterned to give it a characteristic size, shape, and surface impression.
        \item A waffle looks like an egg-shaped pancake with deep indentations.
        \item A waffle looks like a grid of squares.
        \item A waffle is a pancake-like pastry that is made from a leavened batter or dough and is cooked between two heated plates.
        \item A waffle is a pancake-like food that is cooked in a waffle iron.
        \item A waffle is a hotcake with different patterns on it.
    \end{enumerate}}
}

\newpage

\subsection{Generated descriptions for ``avocado'' (frequent)}
\vspace{-25pt}
\texttt{
    {\begin{enumerate}
        \footnotesize
        \setlength{\itemsep}{0pt}%
        \setlength{\parskip}{0pt}%
        \item An avocado looks like a pear-shaped fruit with green or blackish skin.
        \item It is a green fruit that has a dark brown or black seed in the center.
        \item An avocado is a pear-shaped green fruit with smooth, green skin and a large seed in the center.
        \item An avocado is a fruit that is brown and bumpy on the outside and green and creamy on the inside.
        \item An avocado is a fruit with a dark green or blackish skin and a soft, fleshy inside.
        \item An avocado is a green, pear-shaped fruit with a smooth, fleshy texture.
        \item An avocado is a pear-shaped fruit with smooth, green skin.
        \item An avocado is shaped like an egg and has a greenish-brownish skin.
        \item An avocado is typically a dark green or black color on the outside with a soft, light green or yellow color on the inside.
        \item An avocado is a pear-shaped fruit with smooth, green skin and a large, pit in the center.
    \end{enumerate}}
}

\subsection{Generated descriptions for ``sausage'' (frequent)}
\vspace{-25pt}
\texttt{
    {\begin{enumerate}
        \footnotesize
        \setlength{\itemsep}{0pt}%
        \setlength{\parskip}{0pt}%
        \item A sausage is a cylindrical shape made of ground meat, typically pork, and a variety of spices and other ingredients.
        \item A sausage is typically a ground-up mix of pork, beef, or other meats with spices and salt, encased in a thin skin.
        \item A sausage is typically a cylindrical shaped food made from ground meat, spices, and other ingredients.
        \item A sausage is a long, thin, cylindrical piece of meat.
        \item Sausages are typically long, cylindrical shaped foods made from ground meat and spices.
        \item A sausage is a tubular meat product typically made from ground pork, beef, or poultry.
        \item A sausage is a cylindrical casing filled with meat, typically pork, and various herbs and spices.
        \item A sausage looks like a tubular shape made of ground up meat that is usually encased in a thin layer of intestine.
        \item A sausage is a cylindrical casing of meat that is typically filled with ground pork, although many other variations exist.
        \item When cooked, a sausage is typically cylindrical and can vary in length.
    \end{enumerate}}
}

\subsection{Generated descriptions for ``spectacles'' (frequent)}
\vspace{-25pt}
\texttt{
    {\begin{enumerate}
        \footnotesize
        \setlength{\itemsep}{0pt}%
        \setlength{\parskip}{0pt}%
        \item A spectacle is a pair of eyeglasses.
        \item A spectacles is a type of eyewear that consists of a frame that holds two lenses in front of the eyes.
        \item Spectacles are a type of eyewear that helps people see more clearly.
        \item Spectacles are glasses that are worn in order to improve vision.
        \item A spectacles usually refers to a glass or plastic lens worn in front of the eye to correct vision, or protect the eye from debris, dust, wind, etc.
        \item A spectacles is a type of corrective lens used to improve vision.
        \item A spectacle is a lens worn in front of the eye to correct vision, for cosmetic reasons, or to protect the eye.
        \item A spectacles has a frame that goes around your head and two lenses in front of your eyes.
        \item A spectacles has two glass or plastic lenses in metal or plastic frames that rest on the ears.
        \item A pair of spectacles is a frame that holds two eyeglasses lenses in front of a person's eyes.
    \end{enumerate}}
}

\newpage
\section{Example Image Exemplars}\label{sec:add_img_exem}

In this section we show a selection of image exemplars, for LVIS classes,
found using the process described in Section~\ref{sec:img_exem_source}.
We also give the LVIS frequency category --- rare, common, frequent.
For cases where the image exemplar comes from a dataset with bounding boxes
(LVIS or VisualGenome) we show the bounding box in yellow.

\begin{figure}[!ht]
\centering
\scriptsize
\includegraphics[width=1.0\linewidth]{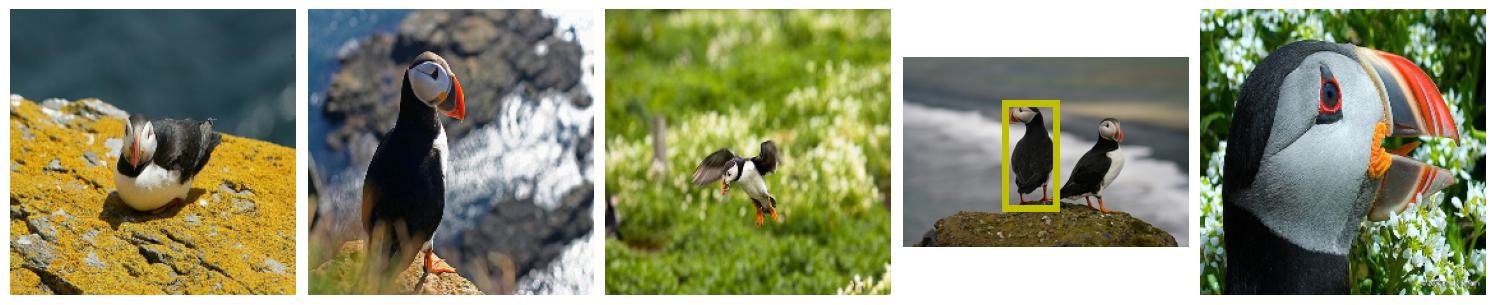}
\vspace{-20pt}
\caption{Image Exemplars for ``puffin'' (rare).}\label{fig:img_exem_puffin}
\end{figure}

\begin{figure}[!ht]
\centering
\scriptsize
\includegraphics[width=1.0\linewidth]{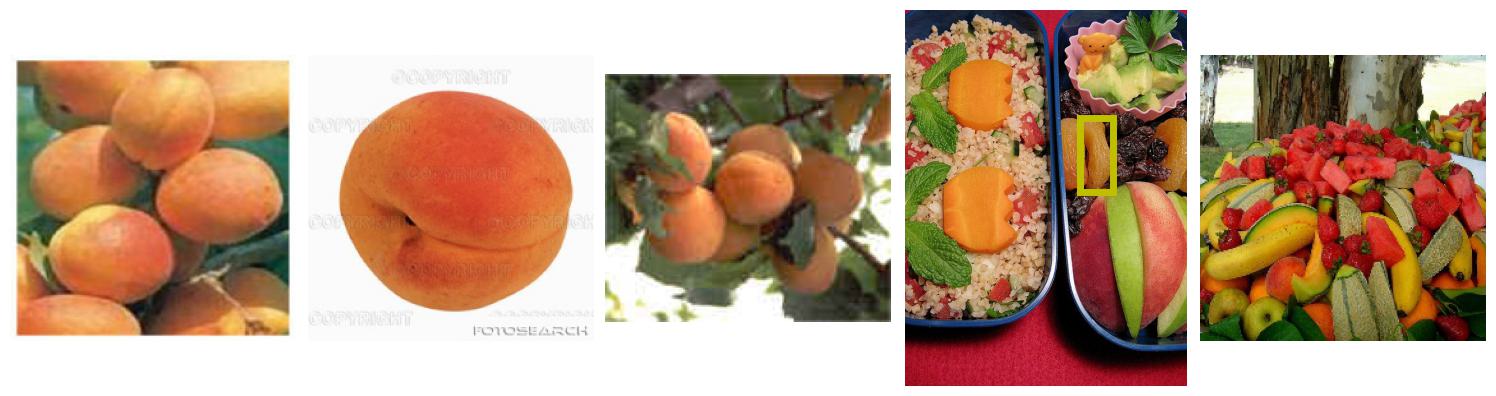}
\vspace{-20pt}
\caption{Image Exemplars for ``apricot'' (rare).}\label{fig:img_exem_apricot}
\end{figure}

\begin{figure}[!ht]
\centering
\scriptsize
\includegraphics[width=1.0\linewidth]{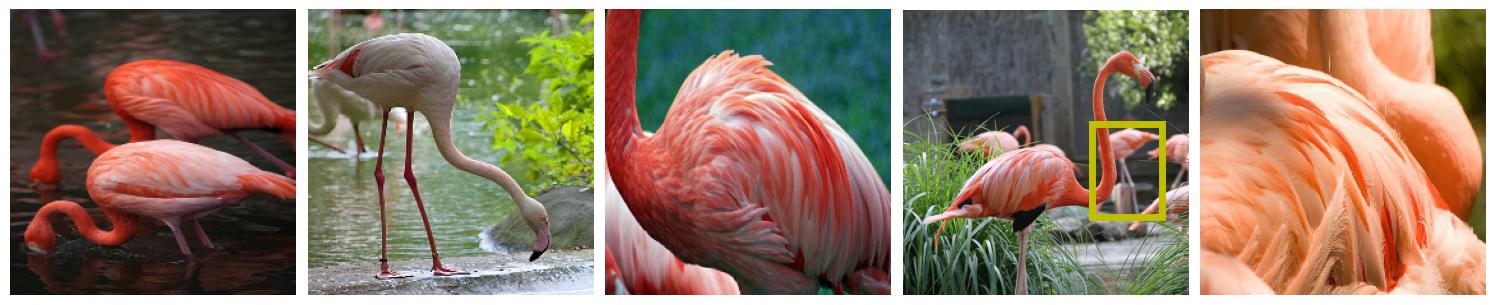}
\vspace{-20pt}
\caption{Image Exemplars for ``flamingo'' (common).}\label{fig:img_exem_flamingo}
\end{figure}

\begin{figure}[!ht]
\centering
\scriptsize
\includegraphics[width=1.0\linewidth]{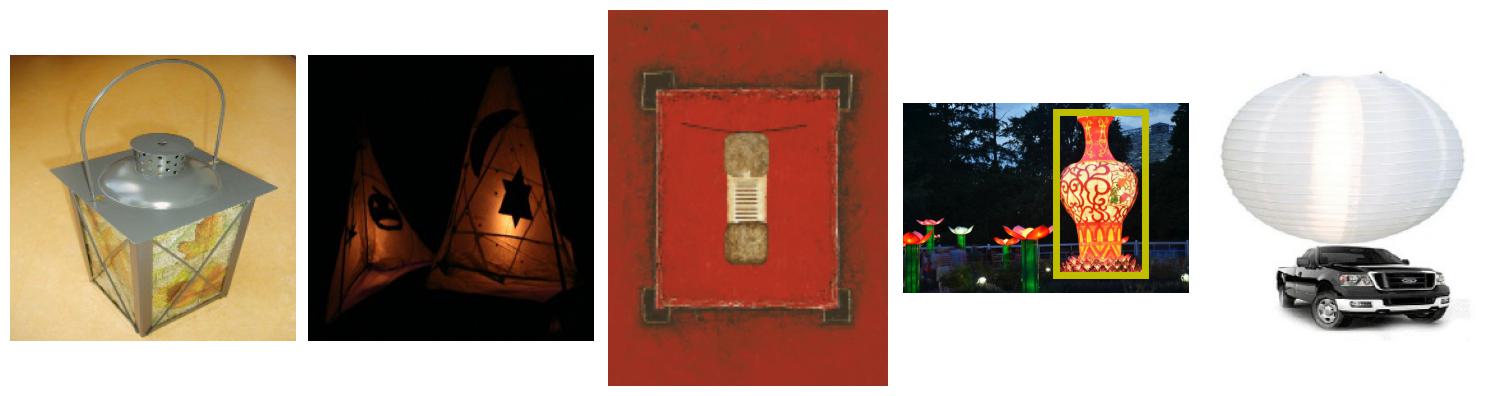}
\vspace{-20pt}
\caption{Image Exemplars for ``lantern'' (common).}\label{fig:img_exem_lantern}
\end{figure}

\begin{figure}[!ht]
\centering
\scriptsize
\includegraphics[width=1.0\linewidth]{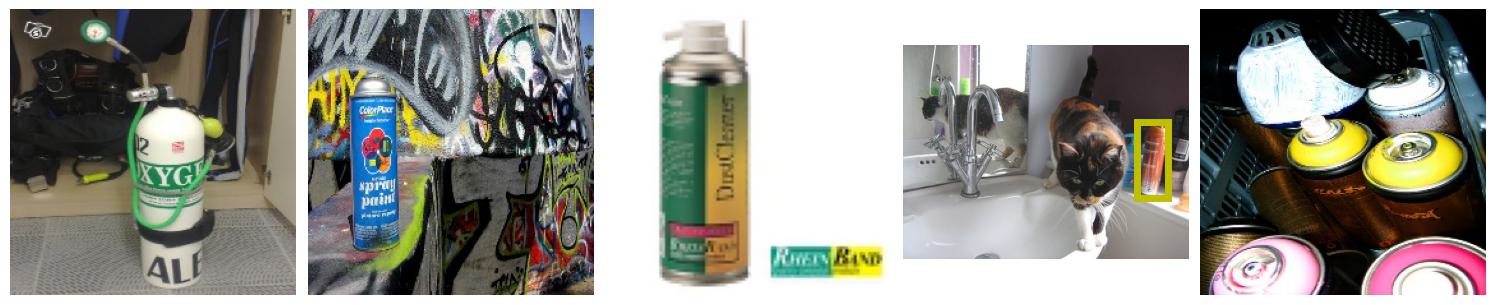}
\vspace{-20pt}
\caption{Image Exemplars for ``aerosol can'' (common).}\label{fig:img_exem_aerosol-can}
\end{figure}

\begin{figure}[!ht]
\centering
\scriptsize
\includegraphics[width=1.0\linewidth]{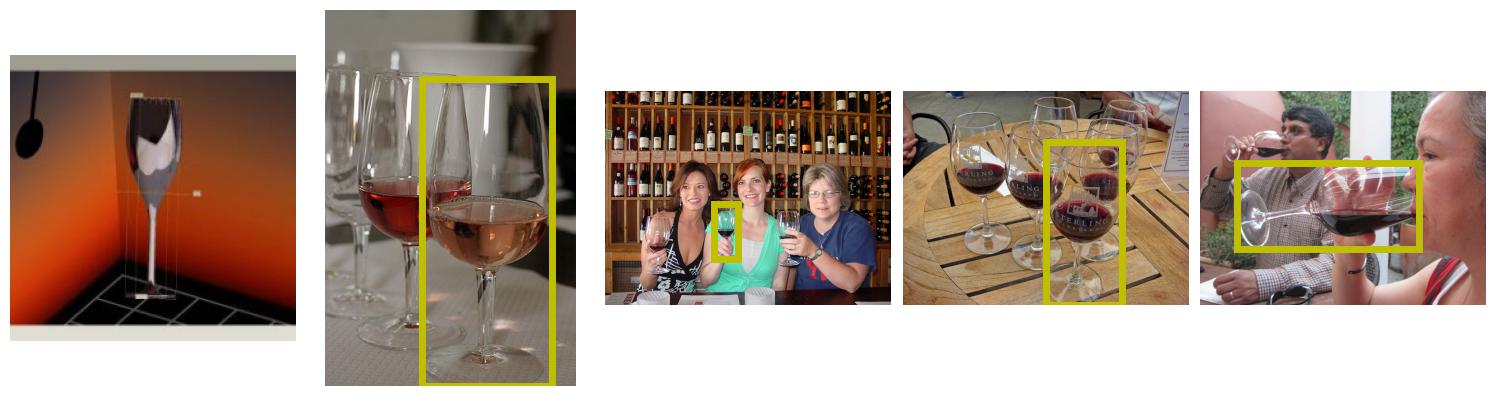}
\vspace{-20pt}
\caption{Image Exemplars for ``wineglass'' (frequent).}\label{fig:img_exem_wineglass}
\end{figure}

\begin{figure}[!ht]
\centering
\scriptsize
\includegraphics[width=1.0\linewidth]{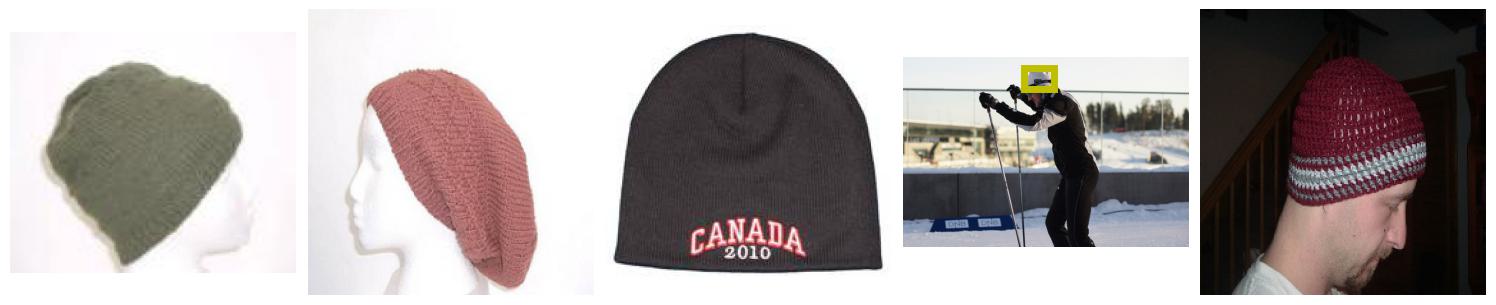}
\vspace{-20pt}
\caption{Image Exemplars for ``beanie'' (frequent).}\label{fig:img_exem_beanie}
\end{figure}

\begin{figure}[!ht]
\centering
\scriptsize
\includegraphics[width=1.0\linewidth]{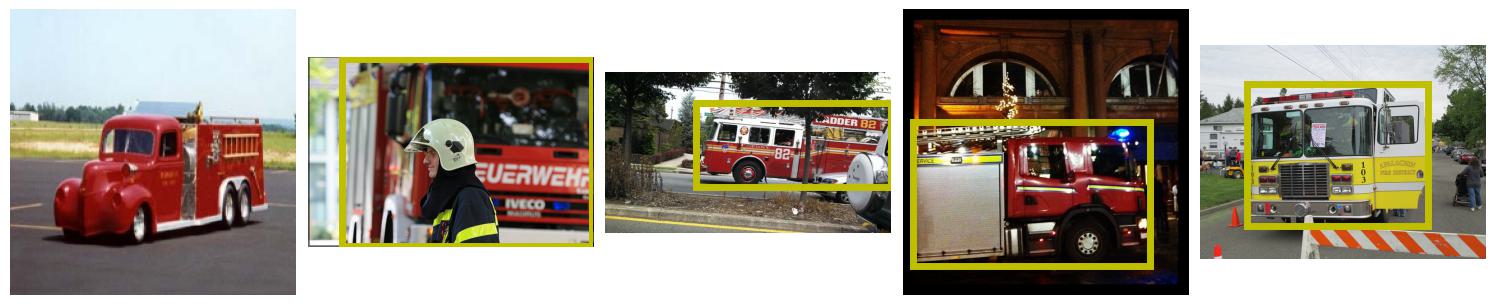}
\vspace{-20pt}
\caption{Image Exemplars for ``fire engine'' (frequent).}\label{fig:img_exem_fire-engine}
\end{figure}

\clearpage

\newpage
\section{More Qualitative Results}\label{sec:qual_res_appen}

In this section we show more rare category detections on the LVIS OVOD benchmark
using our multi-modal classifier trained with IN-L.

\begin{figure}[h!]
\centering
\scriptsize
\includegraphics[width=0.8\linewidth]{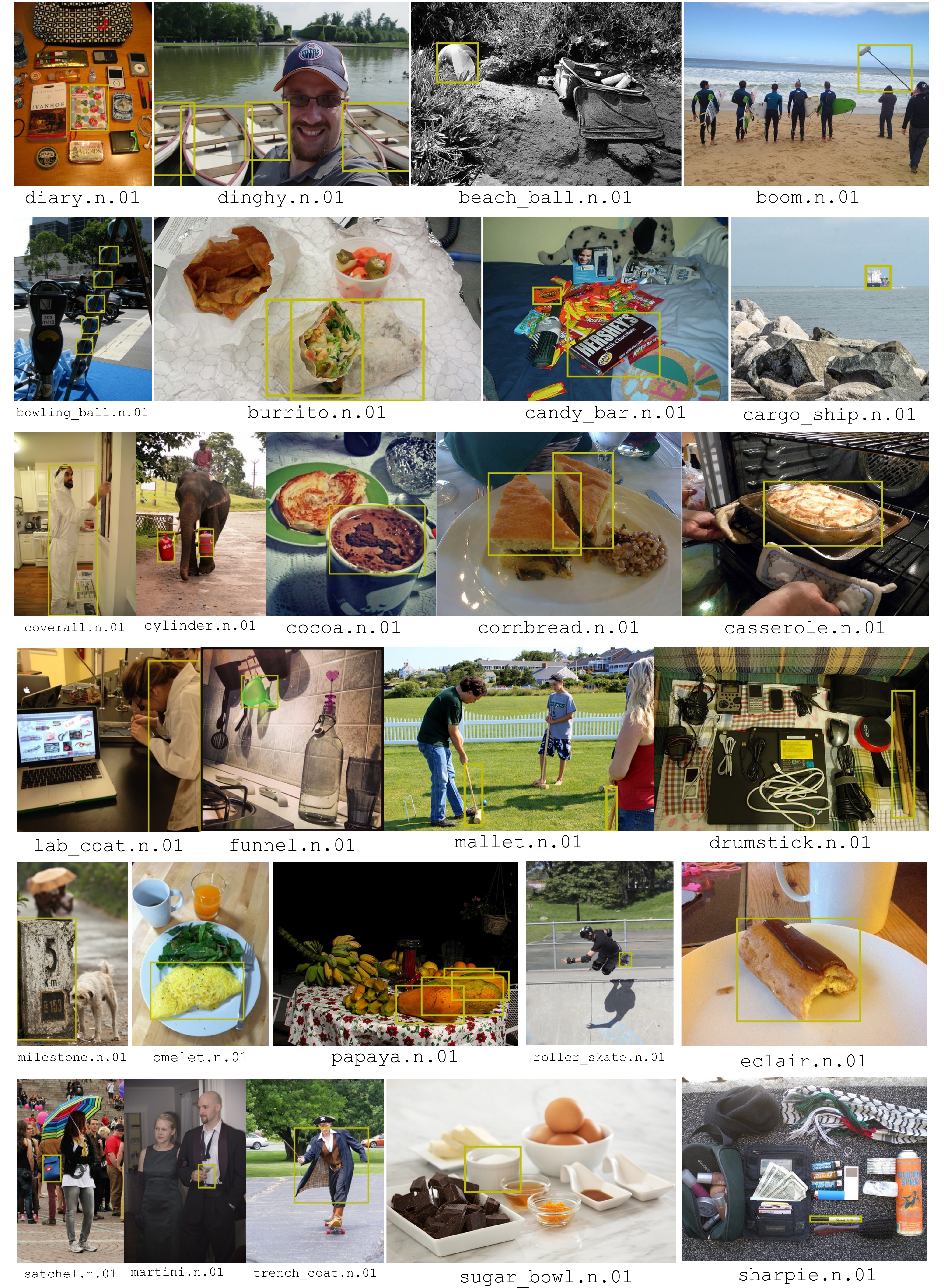}
\caption{Additional qualitative results on LVIS OVOD benchmark.}\label{fig:qual_eg_app}
\end{figure}

\newpage
\section{Dugong}\label{sec:dugong}

\begin{figure}[h!]
\centering
\scriptsize
\includegraphics[width=1.0\linewidth]{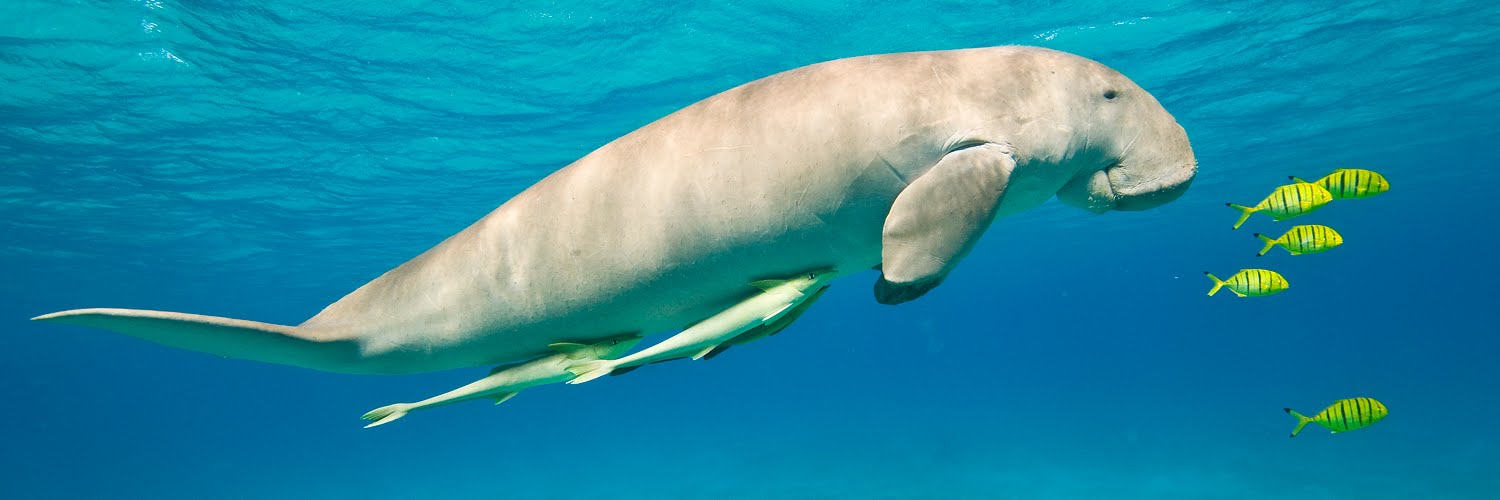}
\vspace{-20pt}
\caption{An example of a dugong --- a visually distinctive marine species with a less
well known name.}\label{fig:img_dugong}
\end{figure}


\end{document}